%% file: main.tex
\pdfoutput=1

\documentclass[11pt]{article}

\usepackage[final]{acl}

\usepackage{times}
\usepackage{latexsym}
\usepackage{booktabs}
\usepackage{xcolor}
\usepackage[utf8]{inputenc}
\usepackage{enumitem} 
\usepackage[T1]{fontenc}

\usepackage[utf8]{inputenc}

\usepackage{microtype}

\usepackage{inconsolata}

\usepackage{graphicx}

\input{header}

\title{\morphogen: A Multilingual Benchmark for Evaluating Gender-Aware Morphological Generation}

\begin{document}
\include{sections/authors}
\maketitle
\begin{abstract}
While multilingual large language models (LLMs) perform well on high-level tasks like translation and question answering, their ability to handle grammatical gender and morphological agreement remains underexplored. In morphologically rich languages, gender influences verb conjugation, pronouns, and even first-person constructions with explicit and implicit mentions to gender. We thus introduce \morphogen\, a morphologically grounded large-scale benchmark dataset for evaluating gender-aware generation in three typologically diverse grammatically gendered languages i.e.\ French, Arabic and Hindi. The core task, \task, requires models to rewrite a first-person sentence in the opposite gender while preserving its meaning and structure. We construct a high-quality synthetic dataset spanning French, Arabic, and Hindi, and benchmark 15 popular multilingual LLMs (2B–70B) on their ability to perform this transformation. Our results reveal gaps and interesting insights into the handling of morphological gender in current models. \morphogen\ offers a focused diagnostic lens for gender-aware language modeling and lays the groundwork for future research on inclusive and morphology-sensitive NLP.
\end{abstract}

\input{sections/introduction}
\input{sections/related_work}

\input{sections/dataset}
\input{sections/experimental_setup}
\input{sections/results_discussion}
\input{sections/Conclusions}
\input{sections/Limitations}
\input{sections/Ethical_Considerations}

\bibliography{custom}

\input{sections/appendix}

\end{document}

%% file: header.tex
\usepackage{graphicx}
\usepackage{subcaption}
\usepackage{url}
\usepackage{latexsym}

\usepackage{amsmath,amssymb}
\usepackage{graphicx}
\usepackage{tabularx}
\usepackage{ragged2e}
\usepackage{multirow}
\usepackage{pifont}
\usepackage{soul}
\usepackage{booktabs}

\usepackage[toc,page]{appendix}
\usepackage{xcolor}
\usepackage{color}
\usepackage{floatrow}
\usepackage{makecell}
\usepackage{xspace}
\usepackage{colortbl}

\usepackage[T1]{fontenc}
\usepackage[utf8]{inputenc}
\usepackage{textcomp}    
\usepackage{tipa}        

\definecolor{salmon}{RGB}{255,121,108}

\usepackage[tableposition=top]{caption}
\usepackage{floatrow}
\usepackage{subcaption}
\usepackage{wrapfig}
\usepackage{makecell}

\newfloatcommand{capbtabbox}{table}[][\FBwidth]

\usepackage{arydshln}
\usepackage{dsfont}

\usepackage{minibox}
\usepackage{tabularx,ragged2e}
\usepackage{tabularx} 
\usepackage{algorithm}
\usepackage[noend]{algpseudocode}
\usepackage{bbm}

\newcommand{\morphogen}{\texttt{MORPHOGEN}}
\newcommand{\dataset}{\texttt{MORPHOGEN}}
\newcommand{\task}{\texttt{GENFORM}}

%% file: sections/authors.tex
\author{
  \parbox{0.9\linewidth}{\centering 
    Mehul Aggarwal$^{\heartsuit}$\thanks{Authors contributed equally}\quad 
    Aditya Agarwal$^{\heartsuit}$\footnotemark[1]\quad 
    Arnav Goel$^{\heartsuit}$\footnotemark[1]\\ 
    Medha Hira$^{\heartsuit}$\footnotemark[1] \quad
    Anubha Gupta$^{\heartsuit}$\thanks{Corresponding author} \\[0.3em]
    $^{\heartsuit}${\rm SBILab, Indraprastha Institute of Information Technology Delhi} \\
    [0.15em]
    {\tt \small anubha@iiitd.ac.in} \\[0.2em]
    \normalsize
    \raisebox{-0.2ex}{\includegraphics[height=0.8em]{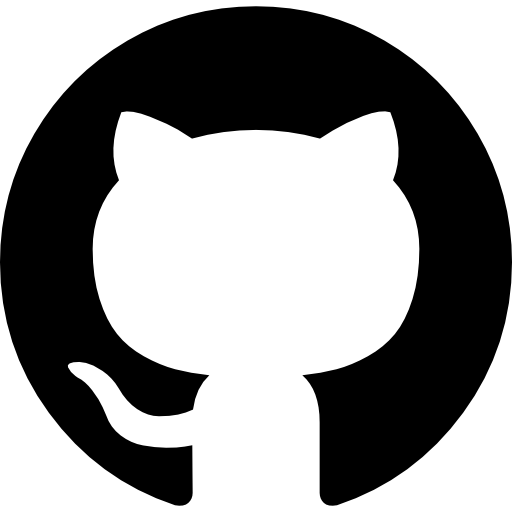}}~
    \href{https://github.com/arnav10goel/Morphogen}{Code}
    \quad
    \raisebox{-0.2ex}{\includegraphics[height=0.8em]{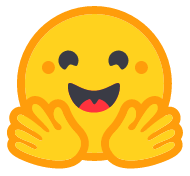}}~
    \href{https://huggingface.co/datasets/ag2003/morphogen}{Dataset}
  }
}


%% file: sections/introduction.tex
\section{Introduction}
\begin{figure*}[!ht]
    \centering  \includegraphics[width=\textwidth]{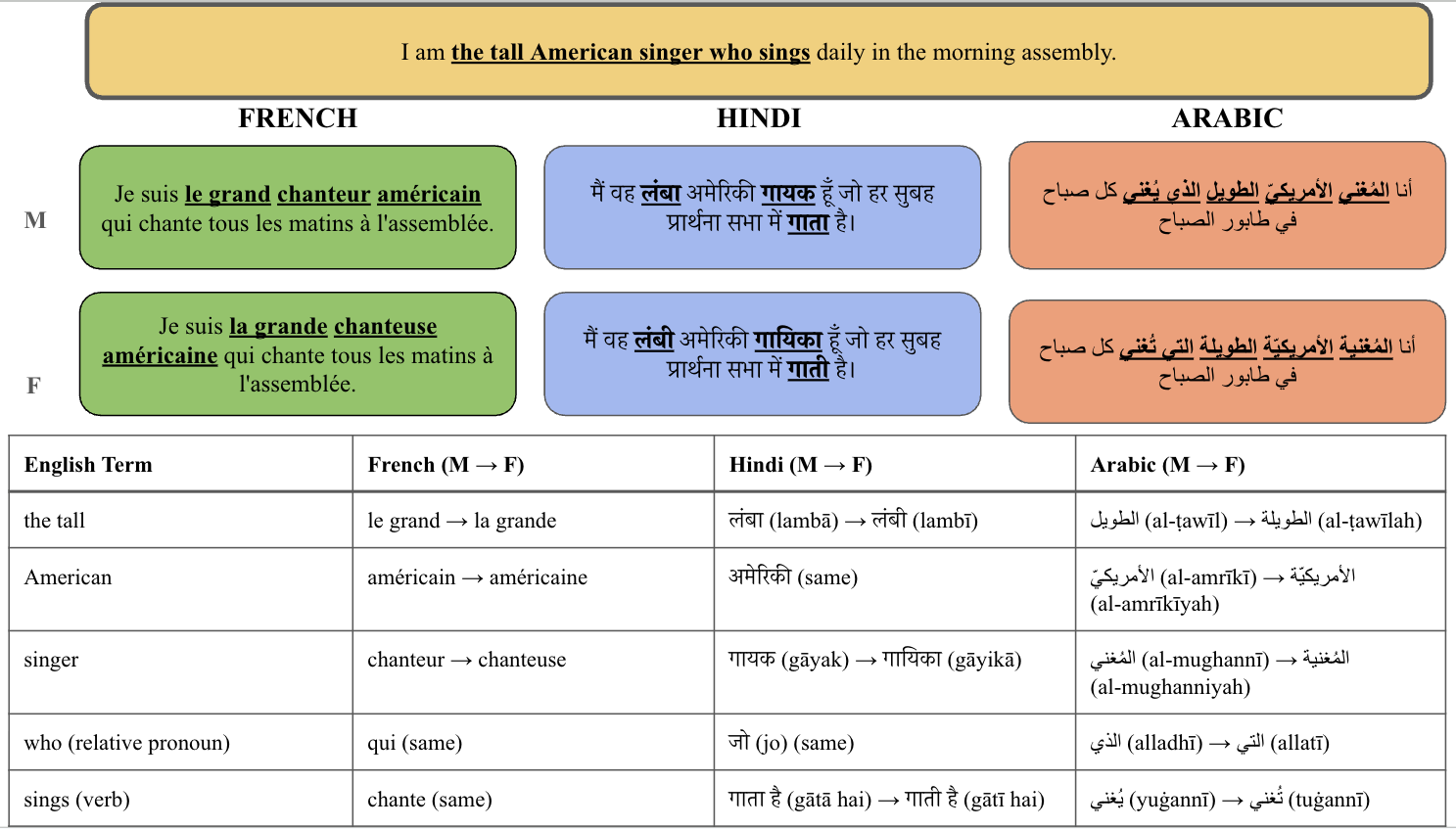}
    \caption{Example illustrating how gender-based morphology differs across the three languages}    \label{fig:gender_explanation_overview}
\end{figure*}

Multilingual large language models (LLMs) demonstrate strong performance across tasks such as summarization, translation, and question answering \citep{goel2023advancementsscientificcontrollabletext, qin2024multilinguallargelanguagemodel, Xu_2025, huang2025surveylargelanguagemodels, 10.1007/978-3-031-49601-1_4, 10.1145/3728483.3760194}. Benchmark datasets like XTREME \citep{hu2020xtreme}, Global-MMLU \citep{singh2024global}, MM-Eval \citep{son2025mmevalmultilingualmetaevaluationbenchmark}, BenchMAX \citep{huang2025benchmaxcomprehensivemultilingualevaluation}, and IndicGenBench \citep{singh-etal-2024-indicgenbench} have become standard tools for evaluating task-specific performance of multilingual LLM models. However, they have been criticized for issues such as poor translation quality, data contamination, and an overwhelming emphasis on high-level tasks that rely heavily on semantic and lexical cues, conflating linguistic competence with broader semantics, making it difficult to isolate fine-grained weaknesses, particularly in morphologically rich or cross-cultural contexts \citep{wu2025bitterlessonlearned2000}.

As LLMs are being increasing deployed across diverse linguistic settings, it becomes essential to evaluate their ability to apply morphological rules in a grammatically coherent manner \citep{piergentili2024enhancinggenderinclusivemachinetranslation, savoldi2025mgentemultilingualresourcegenderneutral}. This is especially critical for languages like French, Arabic, and Hindi, which feature rich grammatical gender constructs, where gender affects verb agreement, pronouns, adjectives, and even word order. For instance, first-person sentences in these languages often contain gendered verbs or adjective forms, even when the subject is implicit (Fig~\ref{fig:gender_explanation_overview}). Gender marking in such cases is morphologically subtle yet semantically significant \citep{gonen2019doesgrammaticalgenderaffect}. Accurate modeling of gender morphology is thus crucial not only for inclusive applications like conversational agents and machine translation, but also for probing how gender bias manifests in LLMs across gendered language structures \citep{sitaram2025multilingual, zhao2024gender, pikuliak2024womenbeautifulmenleaders}.
Despite the linguistic and practical importance of gender morphology, there is currently no benchmark that directly evaluates multilingual LLMs on their ability to reason over and apply gender-specific grammatical rules in syntactically rich constructions. Existing work \citep{joshi2024since, tang2025gendercarecomprehensiveframeworkassessing, sant2024powerpromptsevaluatingmitigating} has primarily tested morphological competence through tokenization or masked word prediction, but falls short of assessing whether models can generate coherent, grammatical sentences conditioned on gender.
To address this gap, we introduce \morphogen\, a morphologically grounded benchmark dataset covering French, Arabic, and Hindi  designed to evaluate gender-conditioned morphological reasoning of LLMs in first-person contexts.

On this benchmark, \textbf{we define the \task\ task as:} given a sentence and the speaker's gender, the model must rewrite the sentence in the opposite gender while preserving grammatical correctness and meaning. To construct linguistically challenging instances, we systematically exploit the rich morphological rules and gender-marking strategies in each language. This requires models to go beyond surface-level transformations and engage in compositional reasoning over linguistic structure. We evaluate \textit{15 widely-used open- and closed-source} multilingual LLMs on this task, spanning model sizes from under \textit{4 billion} to \textit{70 billion parameters}.

Our key contributions are as follows: \textbf{(1)} We present a new benchmark and dataset covering three typologically diverse, grammatically gendered languages: French, Arabic, and Hindi, alongside a parallel English corpus for each sentence (Section~\ref{sec:dataset}). This setup enables evaluation on our proposed task as well as on related NLP tasks such as machine translation and gender bias analysis. To the best of our knowledge, this is the first and most systematically constructed morphology-focused benchmark for these languages, which we plan to publicly release upon acceptance; \textbf{(2)} We introduce novel evaluation metrics to assess the accuracy of gender transformations. These are also applicable to downstream tasks such as translation and gender bias detection in natural language generation (Section~\ref{sec:eval_metrics}); and \textbf{(3)} We benchmark a range of multilingual LLMs on the \task\ task, providing insights into their ability to model and reason about gendered morphological structures.

%% file: sections/related_work.tex
\section{Related Work}

\subsection{Existing Benchmarks on Multilingual LLMs}
Recent advancements in multilingual LLM evaluation have produced several broad-coverage benchmarks. \textbf{XTREME} \citep{hu2020xtreme} emerged as a foundational multi-task benchmark spanning 40 languages and 9 tasks (e.g., NER, QA), though its focus on cross-lingual transfer left gaps in morphosyntactic evaluation. Subsequent works like \textbf{MM-Eval} \citep{son2025mmevalmultilingualmetaevaluationbenchmark} introduced meta-evaluation protocols for 18 languages, emphasizing multilingual consistency in LLM-as-judge scenarios, but remained task-agnostic to gender morphology. Resource-focused frameworks such as \textbf{GlotEval} \citep{luo2025glotevaltestsuitemassively} expanded coverage to hundreds of languages across seven NLP tasks, while \textbf{mHumanEval} \citep{raihan2024mhumaneval} addressed code generation in 200+ languages via machine-translated prompts. Domain-specific efforts like \textbf{MuST-SHE} \citep{bentivogli2020genderdangerevaluatingspeech}. and \textbf{WinoMT} \citep{stanovsky-etal-2019-evaluating} pioneered gender-disambiguated MT datasets for Romance languages, though their narrow scope (1k examples per language) limited utility for LLM evaluation. 

\subsection{Evaluating Gendered Languages in Multilingual LLMs and NLP Systems}

Grammatically gendered languages such as French, Arabic, and Hindi present unique evaluation challenges due to their rich morphological systems. Prior work has shown that large language models (LLMs) often struggle with correctly realizing gender agreement across these languages. For instance, in Hindi, models exhibit errors in gender-inflected verb conjugations and occupational noun morphology \citep{hada2024akalbadiyabias}. In Arabic, evaluations reveal gaps in handling gender agreement across dialectal variations \citep{rhel2025largelanguagemodelsarabic}, while studies in French demonstrate a tendency for models to default to masculine forms despite contextual cues \citep{a1b052d12244474d903c550a762b42e9}.

Despite these findings, recent work \citep{mihaylov2024elegantbridgemultilingualllms} highlights that existing benchmarks do not systematically evaluate the application of gender morphology rules across diverse linguistic typologies. This limitation motivates our work. In contrast to prior datasets like Holistic Bias \citep{smith2022holisticbias}, which focus on English descriptors of gender and identity, our dataset directly targets the morphological realization of gender in multilingual, grammatically gendered settings.

Complementary lines of research further examine how cultural and speech-related factors influence bias in LLMs and NLP systems, underscoring the importance of addressing gender bias through multiple perspectives and modalities \citep{goel2024exploringmultilingualunseenspeaker, li2025attributingcultureconditionedgenerationspretraining, goel2024multilingualprosodytransfercomparing, hira2024crossvoice}.

%% file: sections/dataset.tex
\section{Dataset} 
\vspace{-0.5em}
\label{sec:dataset}
In this section, we introduce the \morphogen\ dataset. We first describe the dataset, the reason for choosing the specific languages, inform  what we mean by gendered terms, and explain the rules. Next, we explain the dataset construction process, compare it to existing parallel corpora, and explain the \task\ task formulation.

\subsection{Dataset Description and Statistics}
Our proposed dataset, \dataset, covers three grammatically gendered languages: French, Arabic, and Hindi. For each language, we construct a corpus of sentence pairs. Each sentence pair exists with the first person speaker as masculine and as feminine, along with a parallel English version. Thus, for each sentence, we have its \textit{gender counterfactual} as a ground truth, which is used to define a \textbf{Gendered Term}. In other words, gendered terms refer to words that differ between a source sentence and its gender counterfactual.
\input{tables/dataset_stats_table}

As shown in Table-1,
the dataset includes 9,999 French, 2719 Arabic, and 7,610 Hindi sentence pairs. Figure~\ref{fig:gendered_terms_dist} illustrates the distribution of gendered terms per sentence pairs, with some containing up to seven gendered elements, highlighting the morphological complexity of our task.

\input{images/gendered_terms_distribution}

\begin{figure}[h]
    \centering
\includegraphics[width=\textwidth]{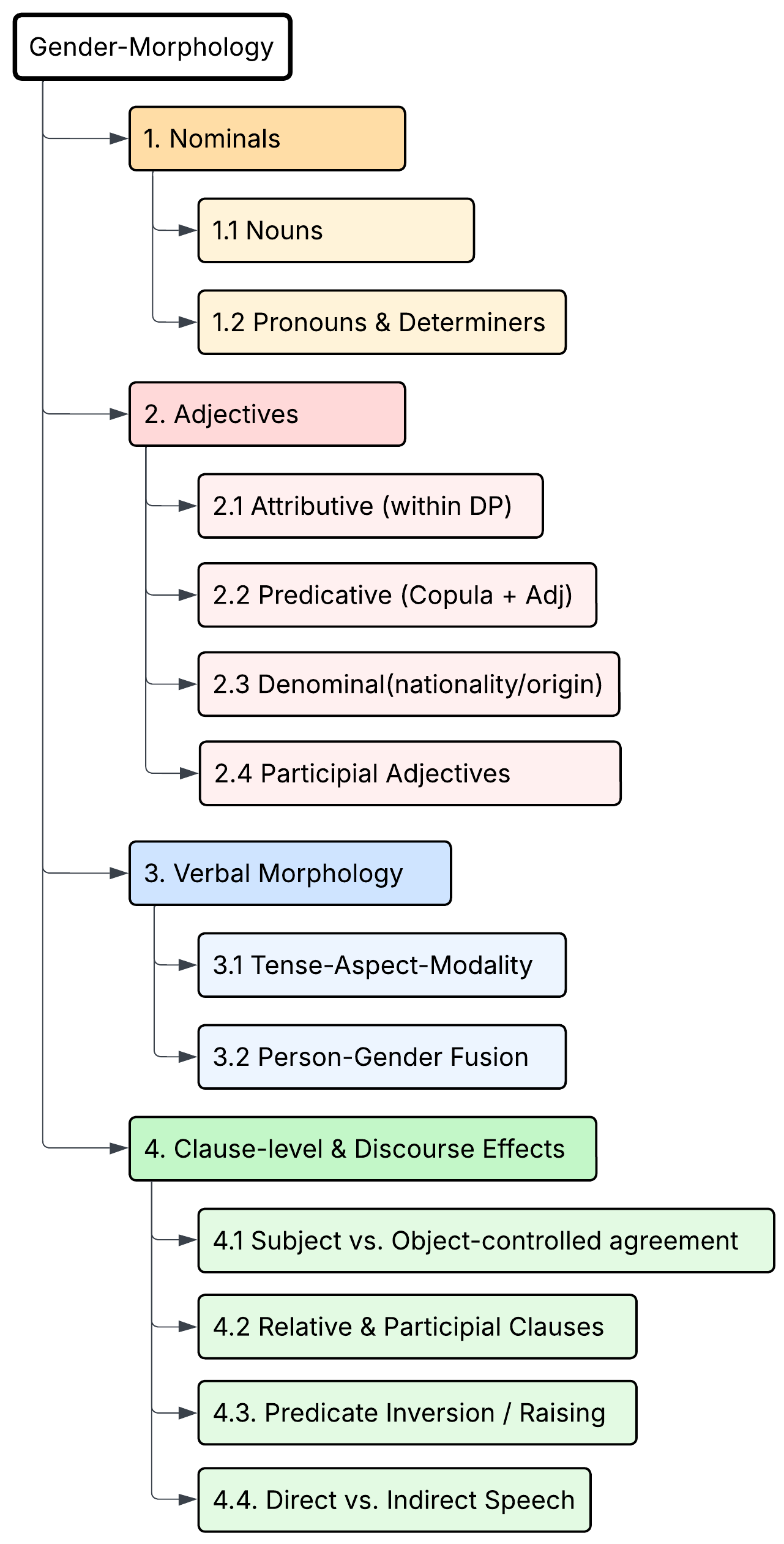}
    \caption{General morphological rules for grammatically gendered languages}
    \label{fig:overview_generic_morph}
\end{figure}

\vspace{-1em}
\subsection{Task Formulation} \label{sec:task_formulation}
For the proposed \task\ task on \morphogen, we prompt a multilingual LLM with a first-person sentence to rewrite the sentence in the opposite gender, i.e., from masculine to feminine or vice versa, based on the original speaker's gender. The model must correctly apply language-specific morphological rules while preserving the sentence’s meaning, fluency, and syntactic structure. 

\subsection{Gender Morphology for Chosen Languages} \label{sec:gender_morph_chosen}
\dataset\ comprises sentence pairs in three typologically diverse, grammatically gendered languages: \textbf{French}, \textbf{Arabic}, and  \textbf{Hindi}. {These were deliberately selected to capture a range of gender assignment strategies of semantic and morphological nature, offering a diverse testbed for evaluating morphological behavior in multilingual LLMs across the selected languages.} All three languages feature binary gender systems (masculine and feminine), but differ significantly in how gender is marked and propagated. This variation is depicted in Figure \ref{fig:gender_explanation_overview}.

\textbf{French} combines semantic, morphological, and phonological cues. While suffixes like \textit{-e} often indicate feminine gender, exceptions are common. Gender agreement is mandatory across determiners, adjectives, and verbs, but variability in marking makes it typologically distinct.

\textbf{Arabic} features a highly regular morphological system where gender is marked primarily via suffixation (e.g., \textit{-a} for feminine). Agreement is strict and pervasive across verbs, adjectives, and pronouns, making it a consistent ground for evaluating morphological accuracy.

\textbf{Hindi} employs a natural gender system with partial morphological marking. Gender is semantically assigned, especially for animate nouns, and commonly marked via suffixes (e.g., \textit{-ā} for masculine, \textit{-ī} for feminine). Agreement extends to verbs, adjectives, and pronouns, but with moderate regularity due to exceptions.

Together, these languages exemplify distinct typological frameworks in gender morphology: French integrates phonological, morphological, and semantic gender assignment; Arabic employs regular morphological suffixation with strict agreement; and Hindi blends semantic natural gender with morphological suffixes. 

\vspace{-0.5em}
\subsection{Construction of Morphological Rules} \label{sec:rule_constr}
To evaluate the performance of multilingual models on gender transformation in first-person contexts, we constructed a set of language-specific morphological rules grounded in linguistic theory (shown in Figure~\ref{fig:gender_rule_dist_overall} of Appendix). These rules are inspired by a general taxonomy of gender morphology across grammatically gendered languages (Figure~\ref{fig:overview_generic_morph}) and are illustrated with concrete examples (Table~\ref{tab:gender_morphology_overview_examples} of Appendix). We present an overview of our motivation behind constructing these rules as:\\
\textbf{(1) Verbs and Tenses.} Gender inflection on verbs depends on both tense and aspect, varying across languages. For instance, French present-tense verbs are gender-invariant, while past participles in compound tenses agree in gender with the subject. Our rules capture such tense-specific patterns.\\
\textbf{(2) Adjectives and Role Nouns.} Adjectives and identity-bearing nouns (e.g., occupations, nationalities) often mark speaker gender morphologically. We design transformation rules to reflect these regular and predictable gendered forms.\\
\textbf{(3) Pronouns and Possessives.} Gender marking in pronouns and possessives is language-dependent. Hindi marks the gender of the possessor, while French and Arabic express gender through grammatical agreement. Our rules reflect these alignment differences.\\
\textbf{(4) Clause-Level Effects.} Gender agreement may be influenced by sentence structure, especially in constructions involving passives or object-fronting. We include rules to account for such syntactic interactions that affect gender realization.\\
\textbf{(5) Multiple Entities and Gender Interference.} To evaluate a model's sensitivity to speaker identity, we introduce sentences with two human referents. Only the speaker’s gender governs agreement, allowing us to test susceptibility to gender interference \citep{lee-etal-2024-fine}.

We provide detailed rules with examples for each language in the following tables in the Appendix: French (Tables  \ref{tab:data_fr1_examples}, \ref{tab:data_fr2_examples}), Arabic (Table \ref{tab:data_arabic_examples}) and Hindi (Tables \ref{tab:data_hi1_examples}, \ref{tab:data_hi2_examples}).

\subsection{Dataset Construction}
We constructed the \dataset\ dataset capturing sentence-level gender transformations in French, Arabic, and Hindi through a structured pipeline grounded in linguistic principles. For each language, we began by identifying grammatical phenomena where a speaker's gender influences agreement or lexical choice, such as in tense and voice (e.g., active/passive), occupations and adjectives, pronouns and possessives, and multi-entity contexts prone to gender interference. {As each language has its own gender-marking system and grammatical structures, we created language-specific templates (e.g., ‘I am a ⟨occupation⟩’) and independently generated English source sentences for each language (i.e., the English inputs are not shared across languages), ensuring structural consistency and systematic coverage across cases.} Prompts specifying the rules, lexical arguments (e.g., occupation = doctor), and discourse contexts (e.g., politics, classroom, therapy) were used to generate English sentences via GPT-4o-mini \citep{hurst2024gpt}.These English sentences were translated into Hindi (using IndicTrans2 and GPT-4o-mini) \citep{gala2023indictrans2, hurst2024gpt}, Arabic (Grok-3)\footnote{\url{https://x.ai/grok}}, and French (NLLB-200) \citep{nllbteam2022languageleftbehindscaling}. 

The dataset was refined by multiple bilingual annotators proficient in English and their respective target languages. Each annotator was randomly assigned a subset of the data and instructed to follow the refinement guidelines provided in the appendix, discarding any sentences that did not comply. Subsequently, each sentence was manually corrected into both masculine and feminine forms by the annotators, strictly adhering to the correction guidelines. For detailed annotator instructions, please refer to Appendix~\ref{appendix:annotator_guidelines}. 

Finally, the validity of the dataset was verified by cross-validation among annotators. Every sentence pair was independently reviewed by two annotators. Two evaluation scores were used for this process: the Data Validation Score, which measures the overall proportion of valid samples, and the Inter-Annotator Agreement Score, which reports the fraction of entries where both annotators agreed on the validity judgment. The detailed validation procedure is provided in the Appendix~\ref{appendix:annotator_guidelines}. Across all three languages, the average Data Validation Score and Inter-Annotator Agreement Score were 0.9705 and 0.9495, respectively. A total of eight annotators in the age group 18–21 participated in this process. Language-wise annotation details are presented in Table~\ref{tab:dataset_stats_additional} in the Appendix. The resulting parallel gender-specific annotations form a high-quality gold-standard set for evaluating the model's sensitivity to morphosyntactic gender variation.

\subsection{Comparison with Existing Datasets}

Standard parallel corpora often default to masculine forms when gender is not explicitly marked. For instance, the EuroParl corpus includes speaker metadata but only 30\% of its sentences are spoken by women, resulting in a male bias~\cite{koehn2005europarl}. Such imbalance limits their suitability for evaluating gender accuracy. Specialized challenge sets exist but fall short for our speaker-gender restoration task:

\textbf{(1) WinoMT} targets occupational stereotypes across languages, including English–Hindi, but relies on rigid templates that models may overfit to~\citep{stanovsky-etal-2019-evaluating}. \textbf{(2) MT-GenEval} improves diversity and realism for English–Hindi but lacks first-person sentences and speaker-gender labels~\citep{currey-etal-2022-mt}. \textbf{(3) MuST-SHE} offers speaker annotations and first-person content, but is not publicly available ~\citep{bentivogli2020genderdangerevaluatingspeech}. \textbf{(4) mGENTE} supports gender-neutral generation across languages~\citep{savoldi2025mgentemultilingualresourcegenderneutral}, but lacks speaker-grounded, first-person constructions.
\textbf{(5) Arabic Parallel Gender Corpus 2.0} provides first- and second-person gendered sentence pairs from English--Arabic OpenSubtitles~\citep{lison2016opensubtitles, alkhalifa2022arabicgendercorpus}, but its coverage is limited to a few recurring gender-marking rules. 

Our dataset captures a broader range of morpho-syntactic phenomena, offering a stronger benchmark across Arabic, Hindi, and French. To the best of our knowledge, no existing dataset:
\begin{enumerate}[nosep,leftmargin=*]
    \item Provides male and female translations for every sentence.
    \item Aligns examples with grammatical triggers for gender inflection.
    \item Ensures balanced ground truth for both genders.
    \item Covers the full spectrum of gender-marking phenomena.
\end{enumerate}

Mining real transcripts is inefficient because most sentences are gender-neutral and only a few cover key structures. In contrast, prompting large language models under controlled templates enables efficient generation of diverse, balanced, and linguistically grounded examples across Hindi, Arabic, and French.

%% file: tables/dataset_stats_table.tex
\begin{table}[!ht]
\centering
\begin{small}
\label{tab:data_stats_table}
\caption{Statistics for \morphogen{} dataset}
\vspace{-1em}
\begin{tabular}{|c|c|c|c|}
\hline
\textbf{Statistics} & \textbf{Arabic} & \textbf{French} & \textbf{Hindi} \\
\hline
\textbf{Unique Sentences}   & 2,719 & 9,999 & 7,610 \\
\textbf{Number of Rules}              & 14    & 12    & 13    \\
\textbf{Avg. Gender Terms*}  & 2.02   & 1.78   & 1.43   \\
\textbf{Max. Gender Terms*}  & 7     & 7     & 7     \\
\textbf{Avg. Word Count*}    & 12.34  & 26.76  & 15.46  \\
\textbf{Max. Word Count*}    & 38    & 67    & 87    \\
\hline
\end{tabular}
\end{small}
\footnotesize{*computed per sentence}
\end{table}

%% file: images/gendered_terms_distribution.tex
\begin{figure}[!ht]
    \centering \includegraphics[width=0.95\columnwidth]{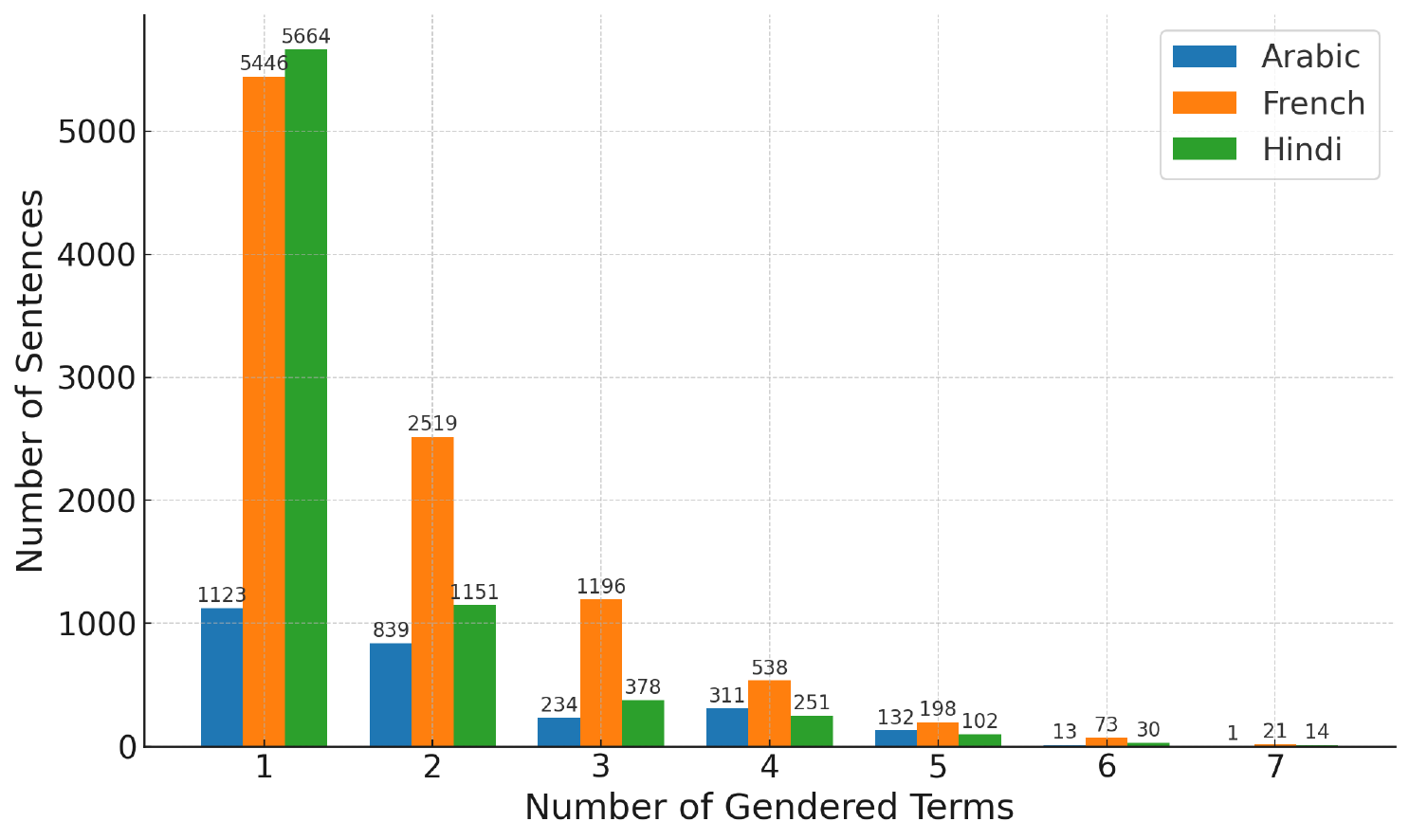}
    \caption{Gendered Terms Distribution in \morphogen}
    \label{fig:gendered_terms_dist}
\end{figure}

%% file: sections/experimental_setup.tex
\section{Experimental Setup}
\subsection{Models Benchmarked}
To effectively evaluate the performance of multilingual LLMs on \morphogen, we conducted extensive benchmarking across 15 models spanning a diverse range of model families and parameter scales. The models evaluated include:
\begin{itemize}[noitemsep, topsep=0pt]
    \item \textbf{LLAMA}: LLAMA-3.1-8B, LLAMA-3.2-3B, LLAMA-3.3-70B \citep{grattafiori2024llama}
    \item \textbf{Qwen}: Qwen3-4B, Qwen3-8B, Qwen3-14B, Qwen3-27B \citep{yang2025qwen3}
    \item \textbf{Gemma}: Gemma2-2B, Gemma2-9B, Gemma3-4B, Gemma3-12B, Gemma3-27B \citep{team2024gemma, team2025gemma}
    \item \textbf{Phi}: Phi4-14B \citep{abdin2024phi}
\end{itemize}

Our goal was to cover a representative and practical spectrum of contemporary multilingual LLMs, ranging from lightweight models (e.g., 2B--4B parameters) suitable for deployment and industry use-cases, to high-capacity models (up to 70B parameters) that are expected to exhibit stronger multilingual generalization. These models were selected based on their widespread adoption, open-source availability, and explicit support for the three gendered languages under study. 

\subsection{Evaluation Metrics} \label{sec:eval_metrics}
To evaluate model performance on the \morphogen\ benchmark, we propose three complementary metrics that measure an LLM's ability to correctly perform gender-aware morphological transformations at different granularities. Note that for any sentence, we collect gendered terms by referring to its gender-counterfactual as presented in Section \ref{sec:dataset}. The proposed metrics are defined as follows:
\paragraph{(1) Sentence-Level Gender Accuracy (SGA):}
This metric measures the proportion of correctly generated gendered terms in each sentence. For a given sentence, we compute the number of gendered words that were correctly modified (i.e., match the gold-standard target) and divide this by the total number of gendered terms in the reference sentence. SGA captures sentence-level precision in handling gendered terms, ensuring correctness at a fine-grained unit of evaluation. The final score is the average of this ratio across all $N$ sentences in the corpora:
\vspace{-0.5em}
\[
\text{SGA} = \frac{1}{N} \sum_{i=1}^{N} \frac{|\text{Gendered}_i \cap \text{Mismatch}_i^c|}{|\text{Gendered}_i|}
\]
As described in Section~\ref{sec:task_formulation}, the \task\ task evaluates bidirectional gender transformation: masculine to feminine and vice versa. We report disaggregated results for each direction, denoted as $\text{SGA}_M$ and $\text{SGA}_F$, corresponding to masculine-to-feminine and feminine-to-masculine conversions, respectively. Additionally, to evaluate any performance gaps between the masculine and feminine disaggregation, we report the gaps between the masculine and feminine scores $\triangle SGA$.
\[
\triangle SGA = SGA_M - SGA_F
\]

\paragraph{(2) Gender IoU Score (GIoU):}
Inspired by the Intersection-over-Union (IoU) metric commonly used in object detection, GIoU metric provides a stricter and more comprehensive measure of morphological transformation quality. It penalizes both over-generation (modifying non-gendered terms or incorrect gendered entities) and under-generation (failing to modify gendered terms). For each sentence, we computed the ratio between the number of correctly transformed gendered terms to the union of gendered and mismatched terms. The final score is the mean of sentence-level IOU values:
\vspace{-0.5em}
\[
\text{GIoU} = \frac{1}{N} \sum_{i=1}^{N} \frac{|\text{Gendered}_i \cap \text{Mismatch}_i^c|}{|\text{Gendered}_i \cup \text{Mismatch}_i|}
\]

This metric captures both precision and recall and is especially useful in sentences with multiple entities or partial gender relevance, where models may hallucinate or overlook certain terms. This metric is particularly useful for evaluating cases of gender interference, where the model incorrectly alters the gender of words associated with entities other than the explicit speaker. In such cases, GIoU penalizes the model for transforming non-gendered terms, thereby ensuring that only valid gender-specific modifications are rewarded. Again, we report disaggregated results for each direction i.e., \textbf{$GIoU_M$} and \textbf{$GIoU_F$}, corresponding to masculine-to-feminine and feminine-to-masculine conversions, respectively. 

\paragraph{(3) Corpus-Level Gender Accuracy (CGA)}
This is a corpus-level aggregation of gender correctness. Instead of averaging per-sentence ratios, we computed the ratio of number of correctly generated gendered terms across the entire test set to the total number of reference gendered terms in the corpus. This provides a holistic measure of overall transformation quality at an \textit{n}-gram level:
\vspace{-0.5em}
\[
\text{CGA} = \frac{\sum_{i=1}^{N} |\text{Gendered}_i \cap \text{Mismatch}_i^c|}{\sum_{i=1}^{N} |\text{Gendered}_i|}
\]

Unlike SGA, which evaluates correctness at the sentence level, CGA extends evaluation to the word level across the entire corpus, which makes it especially effective for longer or more complex sentences with multiple gendered terms.

%% file: sections/results_discussion.tex
\input{tables/results_main_table}
\section{Results and Discussion}
We evaluated 15 widely-used open-source and closed-source multilingual LLMs on the \morphogen\ benchmark across French, Arabic, and Hindi, using the metrics defined in Section \ref{sec:eval_metrics}. The consolidated results are presented in Table \ref{tab:combined_giou_cga_sga}, with detailed language-specific results provided in the Appendix: French in Table \ref{tab:model_results_fr}, Arabic in Table \ref{tab:model_results_arabic}, and Hindi in Table \ref{tab:model_results_hi}. We analyse the variations in performance across different model families and sizes, and discuss the implications for gender bias in these models. 

\vspace{-0.5em}

\subsection{Smaller LMs can’t handle Complex Morphology}
Larger models consistently outperformed smaller ones across all languages, particularly in Arabic, where increased parameter size mitigated morphological complexity. For example, \texttt{Gemma3-27B} (27B parameters) achieved a CGA of 74.74\% in Arabic, markedly outperforming \texttt{Gemma2-2B} at 14.10\%. In Hindi, smaller models remained viable due to simpler rules, with \texttt{LLAMA-3.1-8B} scoring a CGA of 89.21\%, compared to \texttt{LLAMA-3.3-70B} at 91.40\%. French’s larger dataset challenged resource-constrained models, amplifying errors, as \texttt{Gemma2-2B} recorded a CGA of 37.54\%, while \texttt{Phi4-14B} reached 87.70\%. This suggests that parameter size is critical for handling complex morphology but less impactful in simpler linguistic contexts like Hindi.  

\subsection{Masculine Bias in French and Arabic}
Gender bias varied notably across languages, as seen in the $\triangle \text{SGA}$ scores (Figure~\ref{fig:del_sga_all_languages} in Appendix). In Hindi, bias was generally low but occasionally skewed toward feminine forms, with models like \texttt{Gemma3-4B} showing an $\triangle \text{SGA}$ of \texttt{-14.32\%}, often preferring feminine outputs even when the target gender was male. In French, a stronger masculine bias was observed, particularly in larger models such as \texttt{LLaMA3-70B}, which exhibited an $\triangle \text{SGA}$ of \texttt{15.15\%} due to consistent defaulting to masculine forms. Arabic showed persistent masculine bias, especially in plural constructions, with \texttt{Qwen3-32B} recording an $\triangle \text{SGA}$ of \texttt{11.94\%}, frequently generating masculine outputs even in all-female contexts. These trends highlight the influence of gender bias of the training data used in these LLMs and underscore the need for targeted debiasing in morphologically rich languages.  

\subsection{Significant Variance in Model Families}
Architectural differences influenced performance quality. \texttt{Gemma} models excelled in gender fairness, particularly in Arabic, maintaining balance in complex contexts. \texttt{LLAMA} models showed consistency in Hindi and French but struggled with bias in Arabic. \texttt{Qwen} models frequently exhibited masculine bias across languages, suggesting weaker gender handling. \texttt{Phi} models achieved high consistency but faced challenges with entity recognition, especially in Hindi.  

\subsection{Models Misapply Gender in Multi-Entity Sentences}
Gender interference occurs when a model incorrectly alters words associated with all entities’ genders instead of only the gendered terms in sentences with multiple human entities. To measure the correct transformation of gendered terms, we use gender accuracy, which counts only the changes to the intended gendered words. To further penalize any modifications of non-gendered words, we introduce Gendered IoU (GIoU), which is a stricter metric that penalizes models for making unintended edits. These patterns are exemplified through results for the LLaMA family of models on multi-entity cases across languages. Illustrative cases for French, Arabic, and Hindi are presented in Figures~\ref{fig:gender_interference_french}, \ref{fig:gender_interference_arabic}, and \ref{fig:gender_interference_hindi} in the Appendix, respectively. Thus, a large difference between gender accuracy and GIoU indicates that models often transform non-gendered terms and suffer from gender interference and limited instruction following capability for this task.  

\subsection{French: Complex Morphology Amplifies Bias and Challenges Pronoun Agreement}
French’s larger dataset and complex morphology diluted performance, amplifying training imbalances, a trend evident in the GIoU scores presented in Figure~\ref{fig:iou-fr} of appendix. Larger models exhibited masculine bias, while smaller models struggled significantly. Possessive pronoun agreement (e.g., \textit{son instructeur}/\textit{son instructrice}), requiring possession-based gender disambiguation, posed challenges. Smaller models lacked the morphological understanding to handle this, whereas larger models performed more effectively, reflecting the impact of capacity on complex rule application.  

\subsection{Arabic: Lowest Scores with Persistent Masculine Bias in Plurals}
Arabic’s smaller, stricter dataset with intricate morphology yielded the lowest scores, as reflected in the GIoU scores in Figure~\ref{fig:iou-ar} of appendix. Larger models mitigated complexity with balanced gender handling, while smaller models faltered, often showing masculine biases. Female plural agreement (e.g., \textit{ka-mumaththilāt} for ``actresses''), defaulting to masculine for female plural groups, highlighted inadequate training on gender-specific morphology, with most models over-applying masculine forms, even in all-female contexts.  

\subsection{Hindi: Feminine Skew and Entity Errors}
Models achieved higher performance on the Hindi dataset of the \morphogen\ benchmark, reflecting its simpler morphology with fewer gender nuances, as illustrated in the GIoU scores in Figure~\ref{fig:iou-hi} of appendix. Larger models demonstrated superior performance with minimal gender disparity, while smaller models remained competitive, underscoring Hindi’s accessibility. However, some models displayed a feminine bias in female-to-male conversions, and others showed weaker entity recognition due to erroneous gender modifications. Models in 8B--12B range exhibited stronger entity recognition abilities. Smaller models struggled on direct speech involving adjectives and occupations, and co-reference resolution (e.g., \textit{śikṣak}/\textit{śikṣikā} for ``teacher'') failing to resolve a speaker's gender, unlike larger models with robust co-reference handling.  

%% file: tables/results_main_table.tex
\begin{table*}[t]
\small
\centering
\caption{Cross-lingual comparison of Gender IoU (GIoU), Sentence-level Gender Accuracy Gap ($\triangle$SGA), and Corpus-level Gender Accuracy (CGA) across 15 multilingual LLMs for French, Arabic, and Hindi. Higher GIoU and CGA indicate better gender understanding, while lower $\triangle$SGA indicates reduced bias.}
\label{tab:combined_giou_cga_sga}
\begin{tabular}{lccccccccc}
\hline
\textbf{Model} & \multicolumn{3}{c}{\textbf{French}} & \multicolumn{3}{c}{\textbf{Arabic}} & \multicolumn{3}{c}{\textbf{Hindi}} \\
\cmidrule(lr){2-4} \cmidrule(lr){5-7} \cmidrule(lr){8-10}
& \textbf{GIoU ↑} & \textbf{$\triangle$SGA ↓} & \textbf{CGA ↑} 
& \textbf{GIoU ↑} & \textbf{$\triangle$SGA ↓} & \textbf{CGA ↑} 
& \textbf{GIoU ↑} & \textbf{$\triangle$SGA ↓} & \textbf{CGA ↑} \\
\hline
\textsc{Qwen2.5-0.5B} & 5.47 & 4.55 & 4.16 & 4.14 & 8.49 & 4.59 & 0.35 & 0.63 & 0.21 \\
\textsc{Gemma2-2B} & 39.73 & -5.14 & 37.54 & 14.73 & -0.81 & 14.10 & 71.41 & 7.35 & 65.41 \\
\textsc{LLAMA-3.2-3B} & 54.49 & 11.42 & 53.48 & 18.31 & -27.61 & 17.75 & 48.54 & -64.65 & 49.72 \\
\textsc{Gemma3-4B} & 52.70 & -14.16 & 51.60 & 45.68 & -8.20 & 48.93 & 67.50 & -14.32 & 64.58 \\
\textsc{Qwen3-4B} & 58.64 & 7.25 & 53.20 & 34.34 & -0.90 & 35.97 & 62.84 & 3.33 & 68.51 \\
\midrule
\textsc{LLAMA-3.1-8B} & 67.89 & 3.69 & 81.76 & 43.51 & 0.96 & 45.51 & 83.12 & -0.43 & 89.21 \\
\textsc{Qwen3-8B} & 71.66 & 4.86 & 69.91 & 45.89 & 5.03 & 51.01 & 80.96 & 2.16 & 87.82 \\
\textsc{Gemma2-9B} & 60.52 & 1.26 & 55.56 & 46.45 & 2.55 & 45.26 & 85.47 & -7.34 & 84.39 \\
\textsc{Gemma3-12B} & 64.27 & -0.58 & 74.26 & 62.76 & 2.50 & 65.52 & 79.91 & -8.93 & 80.99 \\
\midrule
\textsc{Phi4-14B} & 79.84 & 1.17 & 87.70 & 57.08 & 6.58 & 66.15 & 82.77 & 1.38 & 95.10 \\
\textsc{Qwen3-14B} & 74.22 & 14.23 & 73.91 & 51.83 & 9.45 & 56.08 & 80.68 & 7.81 & 85.80 \\
\textsc{Gemma3-27B} & 71.89 & 7.53 & 79.63 & 70.33 & -0.83 & 74.74 & 77.97 & -7.61 & 82.56 \\
\textsc{Qwen3-32B} & 76.28 & 10.10 & 74.74 & 50.69 & 11.94 & 53.00 & 83.21 & 5.14 & 90.38 \\
\textsc{LLAMA-3.3-70B} & 76.68 & 15.15 & 76.08 & 59.16 & 7.50 & 64.37 & 93.33 & 3.67 & 91.40 \\
\textsc{GPT-4o-mini} & 86.43 & -1.11 & 90.27 & 71.02 & -10.61 & 80.27 & 88.81 & 0.62 & 93.36 \\
\hline
\end{tabular}
\vspace{-0.5em}
\end{table*}

%% file: sections/Conclusions.tex
\vspace{-0.5em}
\section{Conclusions and Future Work}
\vspace{-0.5em}
This paper introduced \morphogen, a new multilingual benchmark for evaluating gender-aware morphological generation in LLMs, covering Hindi, French, and Arabic, three typologically diverse, gendered languages. \morphogen\  focuses on a controlled first-person transformation task that isolates gender-sensitive morphological reasoning. We proposed novel evaluation metrics tailored to this setting and benchmarked 15 multilingual LLMs ranging from 2B to 70B parameters.

Our results show models often confuse gendered forms, especially with multiple entities, and exhibit biased masculine-to-feminine vs. feminine-to-masculine transformations, with some models showing strong directional bias. This highlights persistent limitations in LLMs' handling of gendered morphology.



\morphogen\ offers a foundation for studying morphological competence in multilingual models. Future work should expand it to include 2nd and 3rd person constructions, other gendered languages, and more complex discourse. Our work also enables developing gender-sensitive training and evaluating bias in generative tasks like translation, summarization, and dialogue.

%% file: sections/Limitations.tex
\section{Limitations}
This work presents \morphogen, a large-scale, synthetic benchmark designed to evaluate multilingual language models on grammatical gender and morphological agreement across three typologically diverse and gendered languages: French, Arabic, and Hindi. While we believe \morphogen\ represents an important step toward more inclusive and linguistically grounded evaluation of LLMs, several limitations remain.

\textbf{First}, the dataset currently covers only three languages, each represented in a standardized form without accounting for dialectal variation. {Specifically, we use Modern Standard Hindi, Standard Metropolitan French, and Modern Standard Arabic.} French, Arabic, and Hindi each have dozens of dialects, many of which exhibit distinct grammatical and lexical gender patterns, which are not yet included in this release. \textbf{Second}, our Arabic dataset is smaller than the others, primarily due to limited availability of high-quality source data and fewer native Arabic-speaking annotators. \textbf{Third}, both Hindi and Arabic are predominantly binary-gendered languages; consequently, our current dataset focuses only on male and female speaker forms. We recognize this binary framing as a limitation and aim to extend the dataset to better represent gender as a spectrum in future work. Finally, while we also introduce multi-entity scenarios to evaluate gender interference, these are currently limited to two human referents per sentence. Expanding to more complex discourse scenarios with multiple gendered entities remains an important direction for future research.

Despite these limitations, \morphogen\ provides a valuable and high-precision resource for advancing evaluation of how  of LLMs across linguistically diverse settings.

%% file: sections/Ethical_Considerations.tex
\section{Ethical Considerations}
While \morphogen\ aims to advance fairness and inclusivity by providing a gender-focused benchmark for morphologically rich languages (French, Arabic, and Hindi), we recognize several ethical considerations regarding its development and application.

First, our task formulation currently relies on the binary (masculine and feminine) grammatical categories inherent to these languages, which does not encompass the full spectrum of gender identities. We plan to explore non-binary expansions in future iterations, guided by linguistic feasibility and community consultation. Additionally, to avoid reinforcing cultural or occupational stereotypes, we carefully curated prompts to actively challenge male-default biases (e.g., explicitly using feminine forms for roles like "doctor" or "leader").

Second, we acknowledge the general risk that improved grammatical coherence could be misused to generate harmful text. To mitigate this dual-use concern, \morphogen\ relies exclusively on strictly synthetic prompts and neutral scenarios.

Regarding data creation, annotations were completed by undergraduate students (aged 18–21) who were fairly compensated and certified for their contributions. We prioritized annotator well-being by ensuring all tasks were completely free of sensitive, offensive, or personally identifiable content.

Finally, \morphogen\ is released under a CC BY-NC 4.0 license for research and non-commercial use, with the intent of helping the community build more equitable and linguistically inclusive NLP systems.

\section{Acknowledgments}
We would like to acknowledge the Infosys Center for Artificial Intelligence (CAI) and IIIT-Delhi for their support during this research. We are also deeply grateful to Jagjot Singh, Akshit K Bansal, and Ankit Agarwal for their dedicated assistance in the annotation and creation of \morphogen.

%% file: sections/appendix.tex
\appendix

\input{images/rule_distribution}

\section{Gender-Morphology} \label{appendix:gender_morph}
Different languages express grammatical gender through distinct morphological patterns. An overview of these patterns is shown in Figure~\ref{fig:overview_generic_morph}, with illustrative examples in Table~\ref{tab:gender_morphology_overview_examples}. These patterns motivate our focus on three gendered languages: French, Arabic, and Hindi.

{To support comparisons made in the results section, we define morphological complexity in terms of (i) the number of agreement targets (e.g., verbs, adjectives, determiners), (ii) the regularity vs. irregularity of gender marking, and (iii) the extent to which gender realization depends on syntactic context (e.g., tense, clause structure, or discourse configuration). Under this definition, French and Arabic both exhibit high morphological complexity, but for different reasons. French shows irregular and context-dependent agreement (e.g., gender agreement in past participles but not in present tense), while Arabic exhibits a more systematic yet nuanced morphology with pervasive agreement across verbs, adjectives, pronouns, and constructions such as relative clauses, conditionals, and multi-entity contexts. In contrast, Hindi follows a comparatively more regular and semantically grounded system, with fewer context-dependent variations.}

For each of these languages, we provide example snippets along with the corresponding morphological rules in Tables~\ref{tab:data_fr1_examples}, \ref{tab:data_fr2_examples}, \ref{tab:data_arabic_examples}, \ref{tab:data_hi1_examples}, and \ref{tab:data_hi2_examples}.

\section{Annotator Guidelines} \label{appendix:annotator_guidelines}

\subsection*{Guidelines for Dataset Refinement}
\begin{itemize}[leftmargin=*]
    \item \textbf{Off-Limits Language:} No sentence may contain profanity, hate speech, slurs, or any other abusive or objectionable content. Annotators must ensure compliance with content-policy restrictions.
    \item \textbf{Naturalness:} Sentences should reflect standard conversational phrasing that a fluent speaker would naturally use, avoiding stilted or machine-generated constructions.
    \item \textbf{Uniqueness:} Identical or trivially paraphrased sentences are to be rejected and regenerated.
    \item \textbf{Template Fidelity:} Sentences must follow the syntactic template exactly, without missing slots, extra words, or rearrangements.
    \item \textbf{Domain Coverage:} For every template, sentences must span all conversational domains specified (e.g., academic, healthcare, legal), ensuring diversity.
    \item \textbf{Gender Specificity:} Each English sentence must be designed such that its translations differ between masculine and feminine forms in the target language.
\end{itemize}

\subsection*{Guidelines for Dataset Correction}
\begin{itemize}[leftmargin=*]
    \item \textbf{Fidelity \& Fluency:} Translations must preserve meaning, tone, and register while being grammatically correct and idiomatic in the target language. Annotators should check word choice, tense, punctuation, and readability.  
    \item \textbf{Speaker-Gender Agreement:} All gender-dependent morphology tied to the speaker (verbs, adjectives, pronouns, etc.) must appear in the correct masculine form in the “male” version and the correct feminine form in the “female” version.  
    \item \textbf{Consistency for Implicit Gender Entities:} Gendered terms referring to non-speaker entities must remain identical across male and female translations. For instance, if \textit{friend} is rendered in masculine form in the male version, it must remain masculine in the female version as well.  
\end{itemize}

\subsection*{Dataset Validation Process}

Each data sample was independently assigned a validity score of 1 or 0 by two annotator, indicating full compliance or non-compliance with the annotation guidelines, respectively. The Data Validation Score (DVS) and Inter-Annotator Agreement (IAA) were computed as follows:

\begin{equation}
\text{DVS} = \frac{\sum_{i=1}^{N} \left( s_{i1} + s_{i2} \right)}{2N}
\end{equation}

\begin{equation}
\text{IAA} = \frac{1}{N} \sum_{i=1}^{N} \mathbb{I}\left( s_{i1} = s_{i2} \right)
\end{equation}

where:
\begin{itemize}
    \item $N$ denotes the total number of sentence pairs in the dataset.
    \item $s_{i1}$ and $s_{i2}$ represent the binary validity scores ($0$ or $1$) assigned to the $i^{\text{th}}$ sentence pair by Annotator~1 and Annotator~2, respectively.
    \item $\mathbb{I}(\cdot)$ is the indicator function, returning $1$ if the condition inside is true and $0$ otherwise.
\end{itemize}

The \textit{Data Validation Score} measures the overall proportion of valid samples across all annotators, while the \textit{Inter-Annotator Agreement} quantifies the fraction of samples for which both annotators assigned identical scores.

\begin{table*}[htbp]
  \centering
  \includegraphics[width=\textwidth]{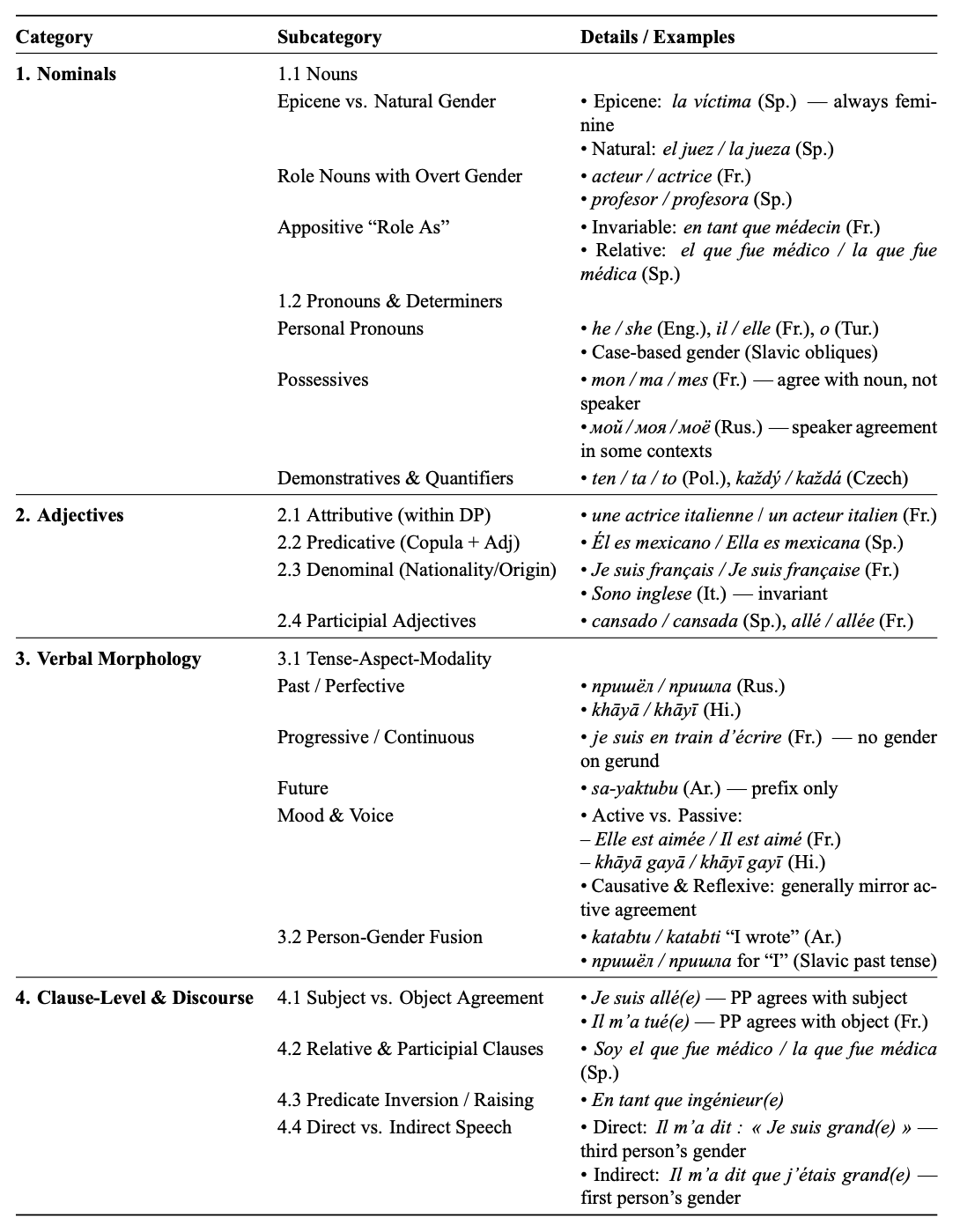}
   \caption{Gender Morphology Overview}
  \label{tab:gender_morphology_overview_examples}
\end{table*}
\begin{table*}[htbp]
  \centering
  \includegraphics[width=\textwidth]{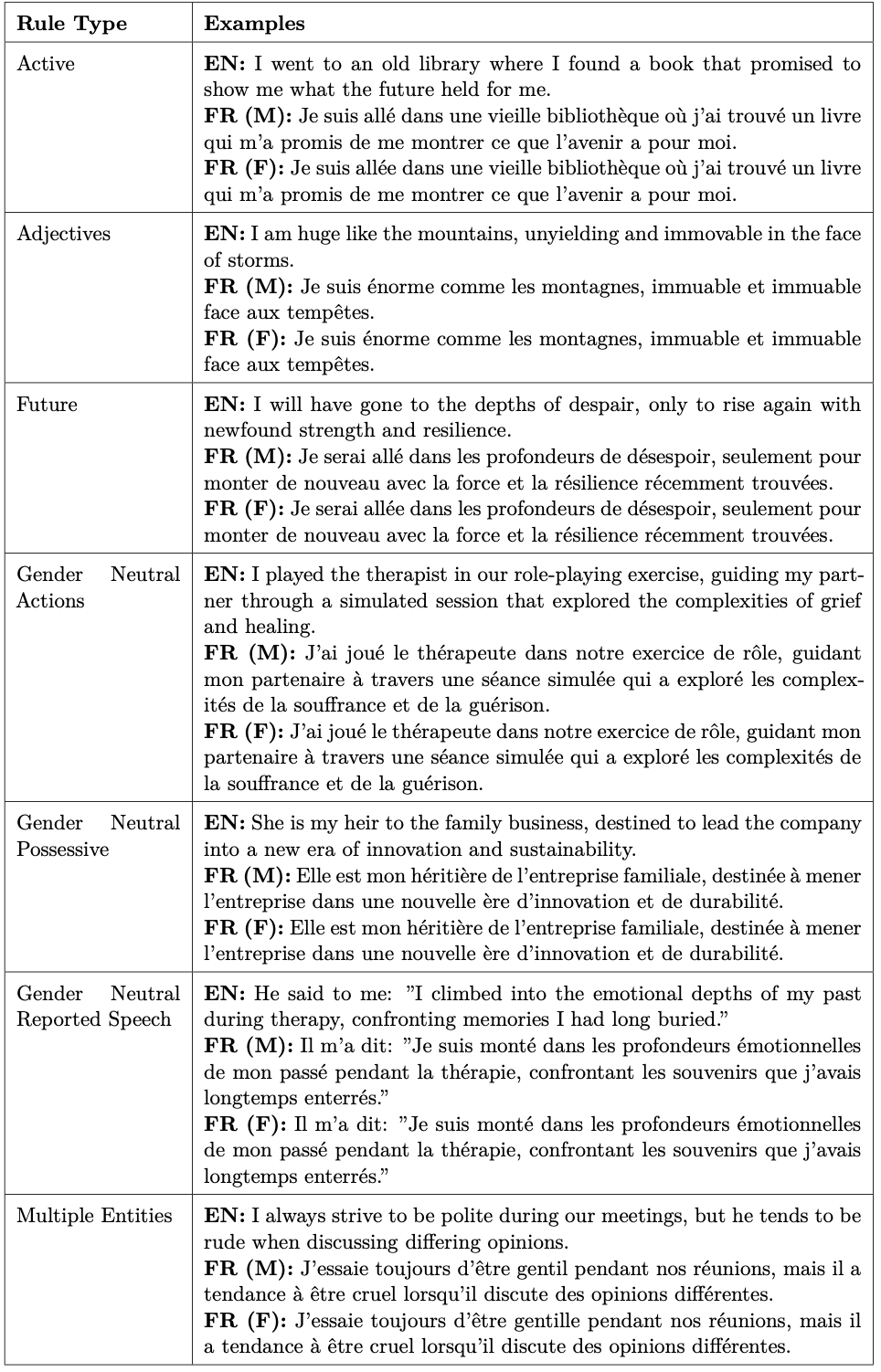}
  \caption{French Gendered Grammar Examples Across Rule Types [1]
}
  \label{tab:data_fr1_examples}
\end{table*}

\begin{table*}[htbp]
  \centering
  \includegraphics[width=\textwidth]{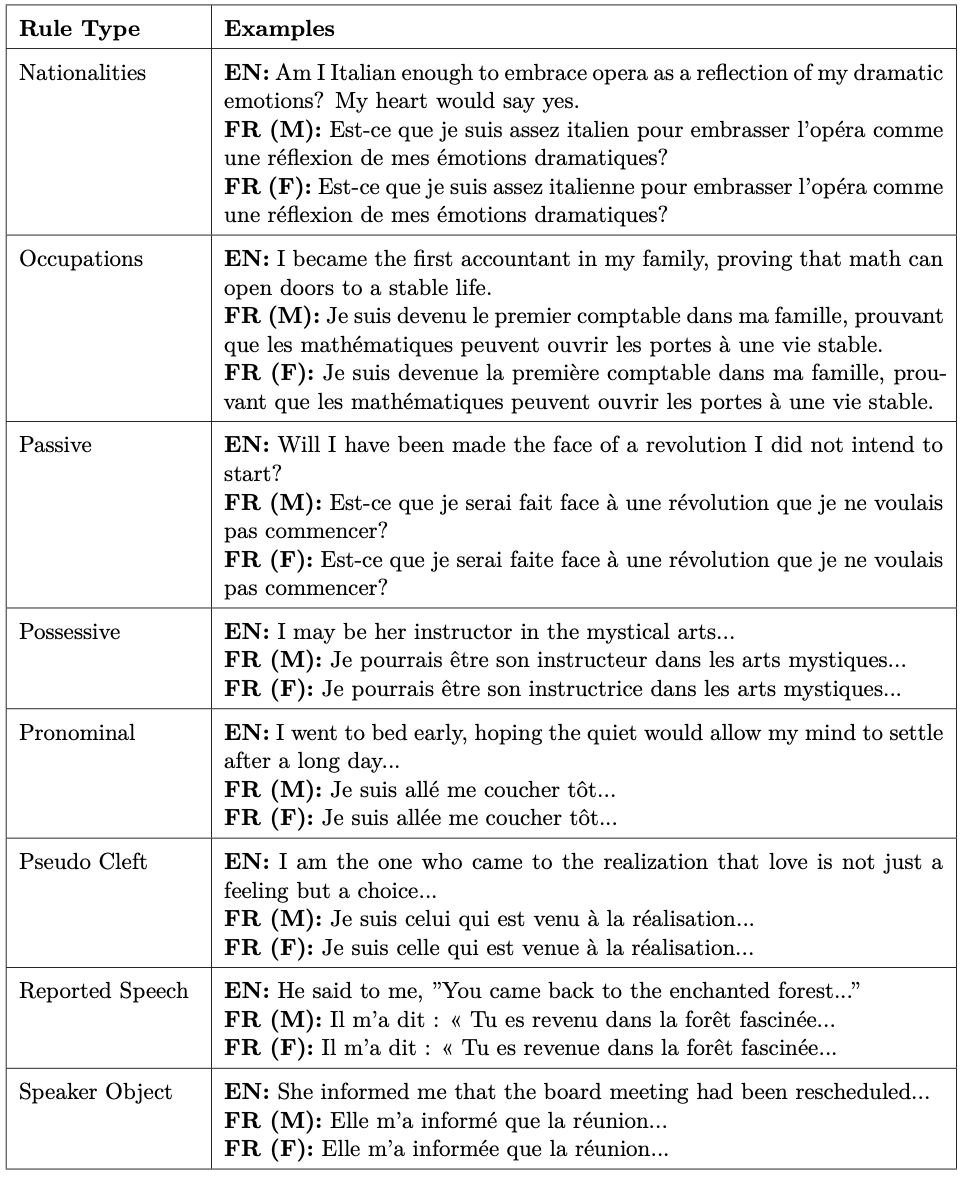}
  \caption{French Gendered Grammar Examples Across Rule Types [2]}
  \label{tab:data_fr2_examples}
\end{table*}
\begin{table*}[htbp]
  \centering
  \includegraphics[width=\textwidth]{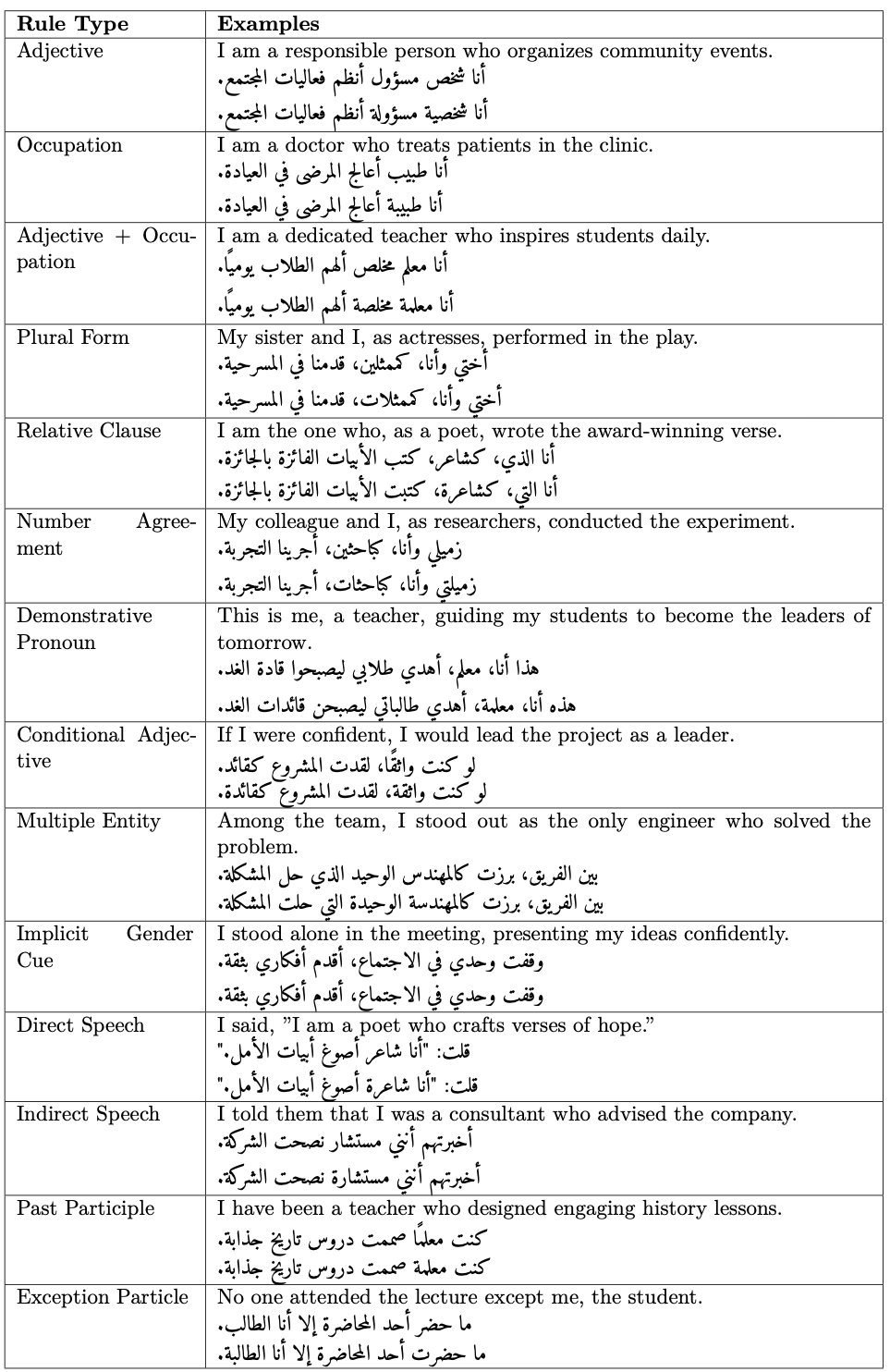}
  \caption{Arabic Gendered Grammar Examples Across Rule Types
}
  \label{tab:data_arabic_examples}
\end{table*}
\begin{table*}[htbp]
  \centering
  \includegraphics[width=\textwidth]{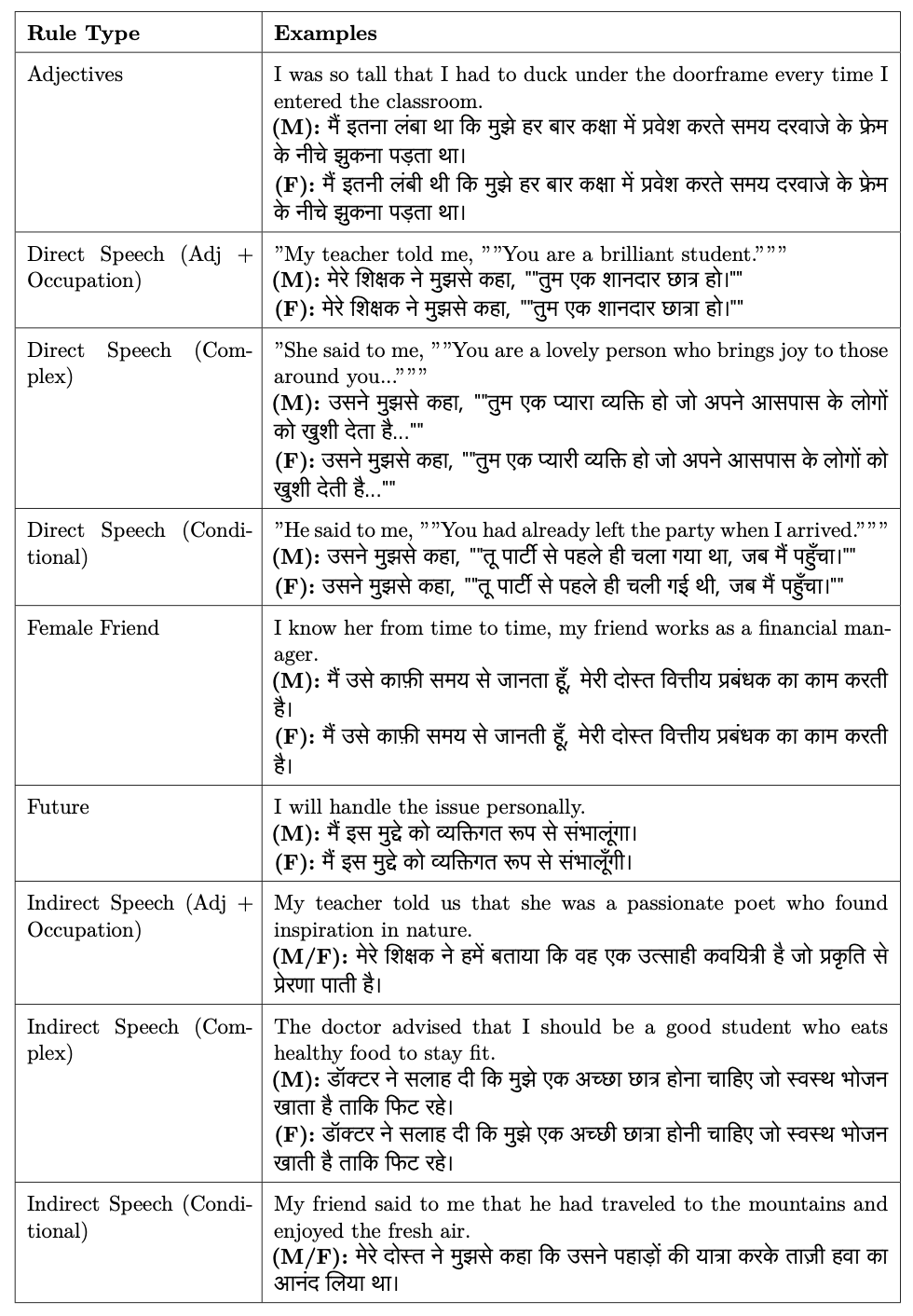}
  \caption{Hindi Gendered Grammar Examples Across Rule Types [1]
}
  \label{tab:data_hi1_examples}
\end{table*}

\begin{table*}[htbp]
  \centering
  \includegraphics[width=\textwidth]{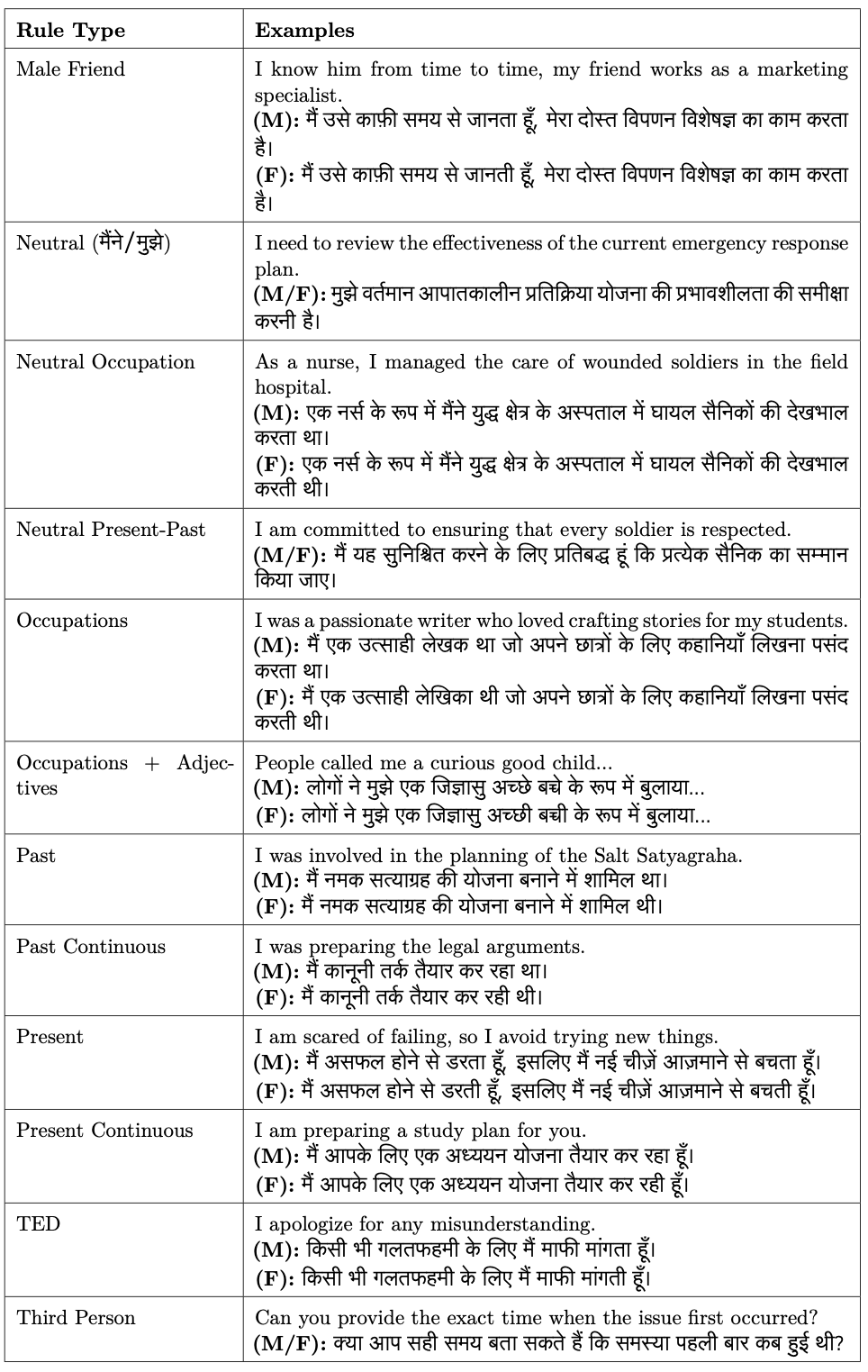}
  \caption{Hindi Gendered Grammar Examples Across Rule Types [2]
}
  \label{tab:data_hi2_examples}
\end{table*}

\begin{table*}[htbp]
  \centering
  \input{tables/kappa_stats}
  \caption{Dataset Statistics Across Languages}
  \label{tab:dataset_stats_additional}
\end{table*}

\input{tables/table_fr} 
\input{tables/table_arabic} 
\input{tables/table_hi} 

\begin{figure*}[t]
    \centering
    \includegraphics[width=0.9\textwidth]{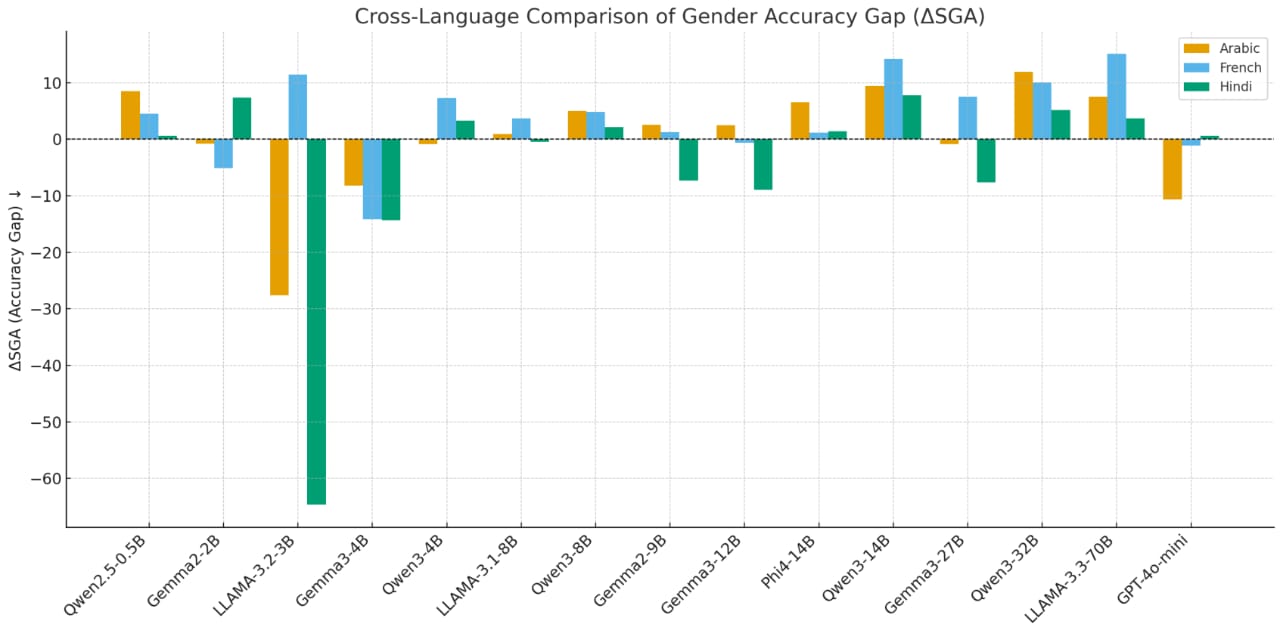}
    \caption{$\triangle \text{SGA}$ (Accuracy Gap) across all models and languages (French, Arabic, Hindi) in the \morphogen\ benchmark. Positive values indicate masculine bias, while negative values indicate feminine bias.}
    \label{fig:del_sga_all_languages}
\end{figure*}




\begin{figure*}[htbp]
    \centering
    \begin{subfigure}[b]{0.4\textwidth}
        \centering
        \includegraphics[width=\linewidth]{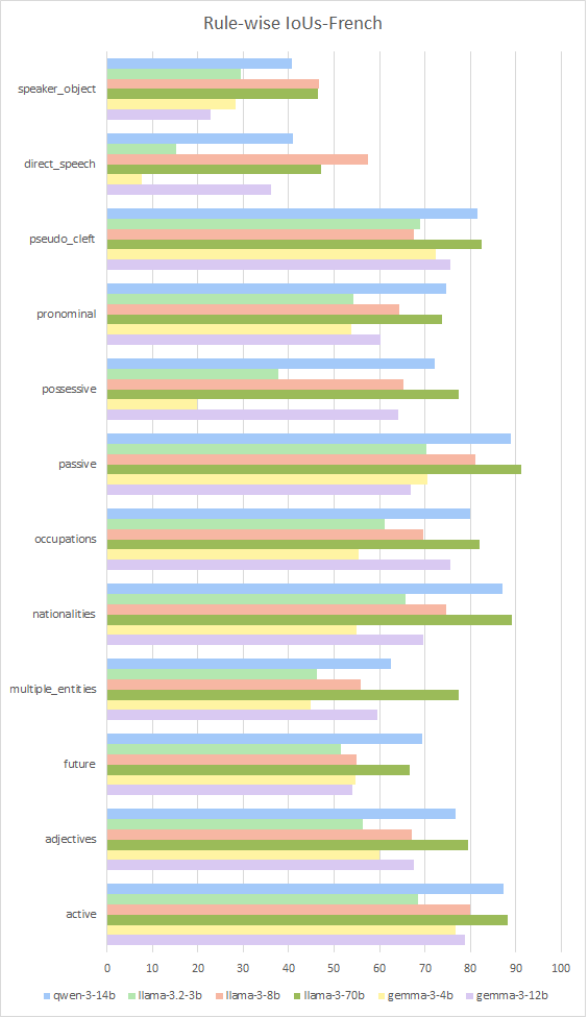}
        \caption{French}
        \label{fig:iou-fr}
    \end{subfigure}
    \hfill
    \begin{subfigure}[b]{0.4\textwidth}
        \centering
        \includegraphics[width=\linewidth]{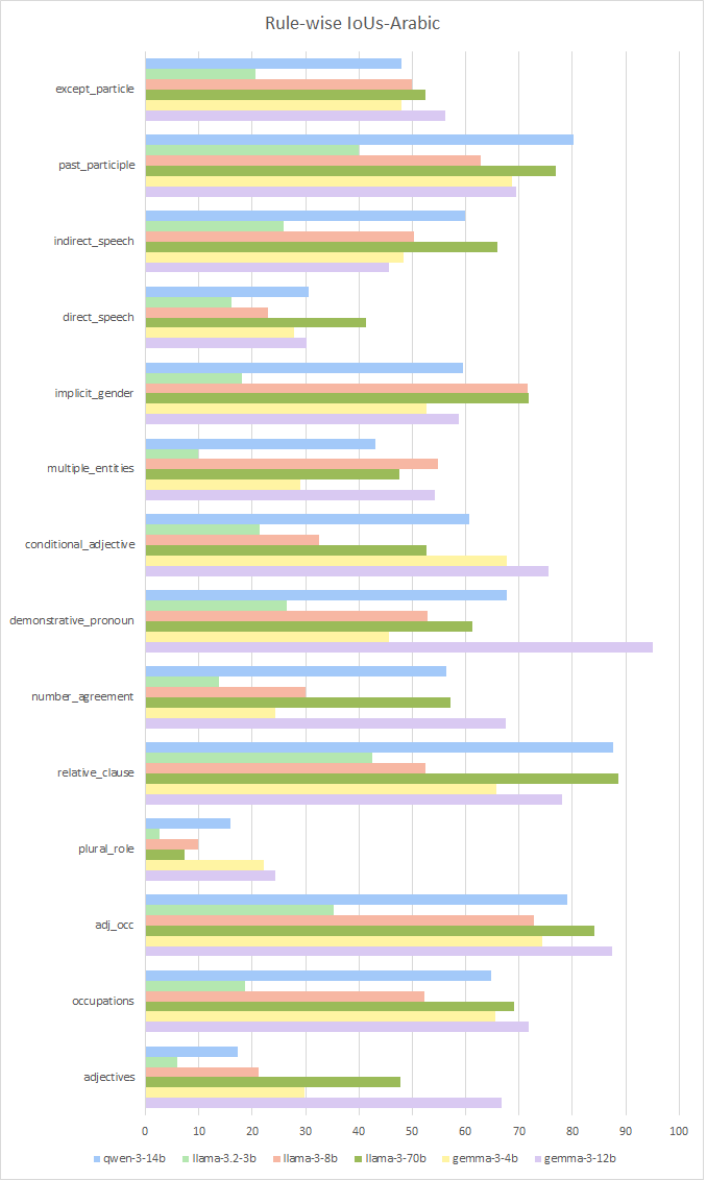}
        \caption{Arabic}
        \label{fig:iou-ar}
    \end{subfigure}
    
    \vspace{1.5em} 
    
    \begin{subfigure}[b]{0.4\textwidth}
        \centering
        \includegraphics[width=\linewidth]{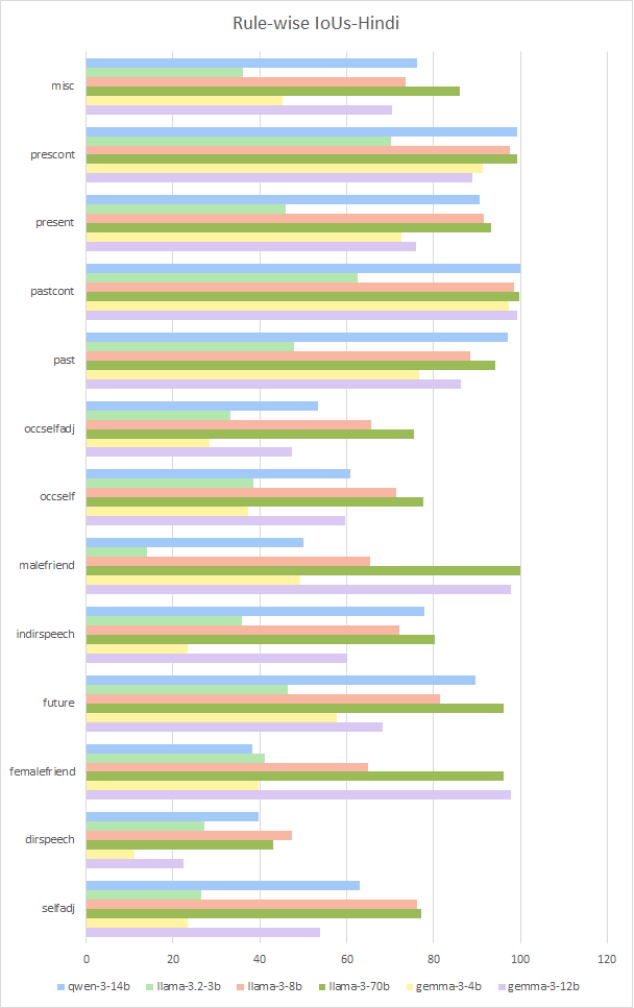}
        \caption{Hindi}
        \label{fig:iou-hi}
    \end{subfigure}
    
    \caption{Rule-based and model-wise IoU metrics across all three languages.}
    \label{fig:iou-metrics-all}
\end{figure*}

\begin{figure*}[htbp]
    \centering
    \includegraphics[width=\textwidth]{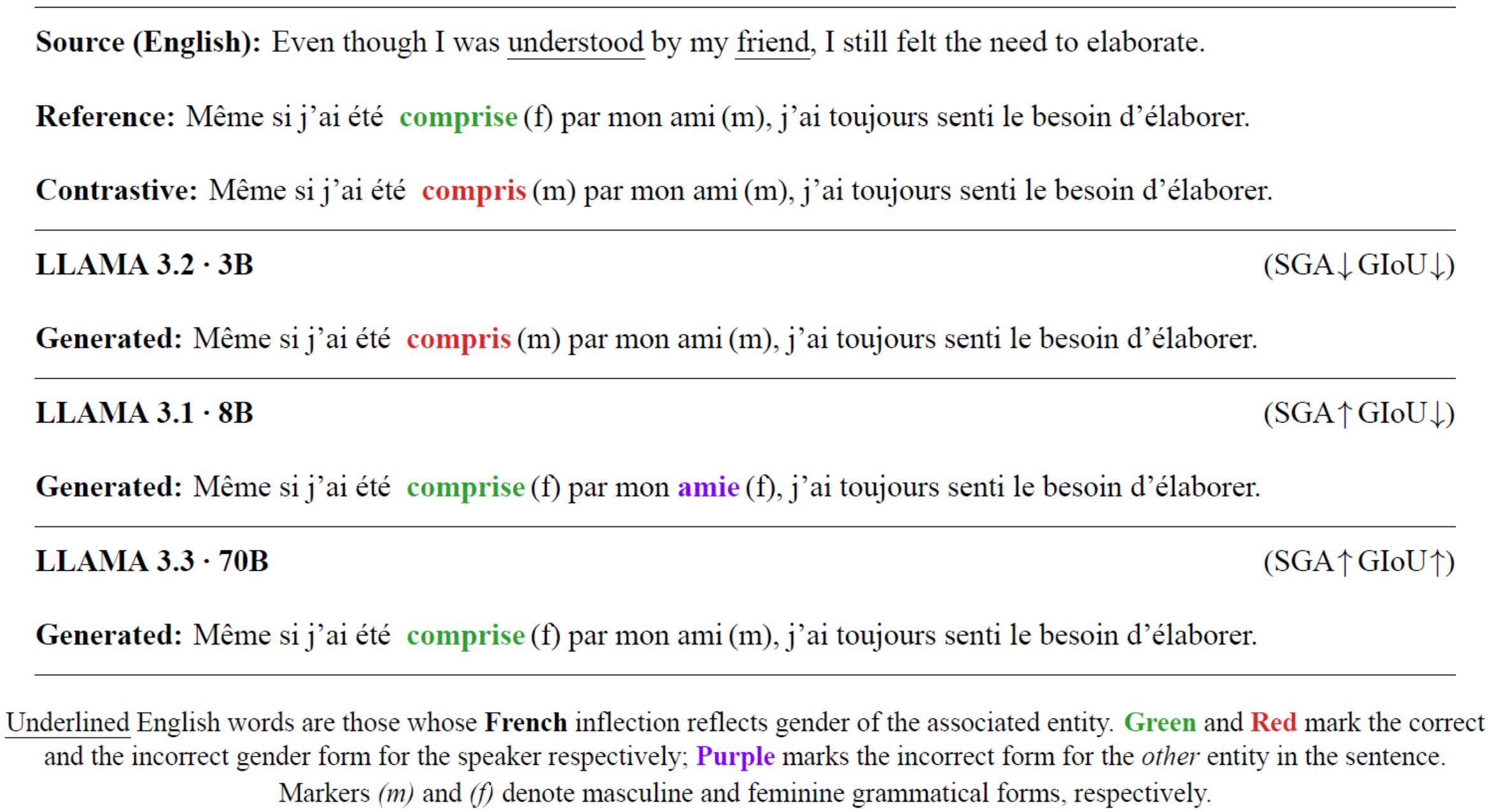}
    \caption{Example of results of \textsc{LLAMA} family of models on multiple entities in French dataset.}
    \label{fig:gender_interference_french}
\end{figure*}

\begin{figure*}[htbp]
    \centering
    \includegraphics[width=\textwidth]{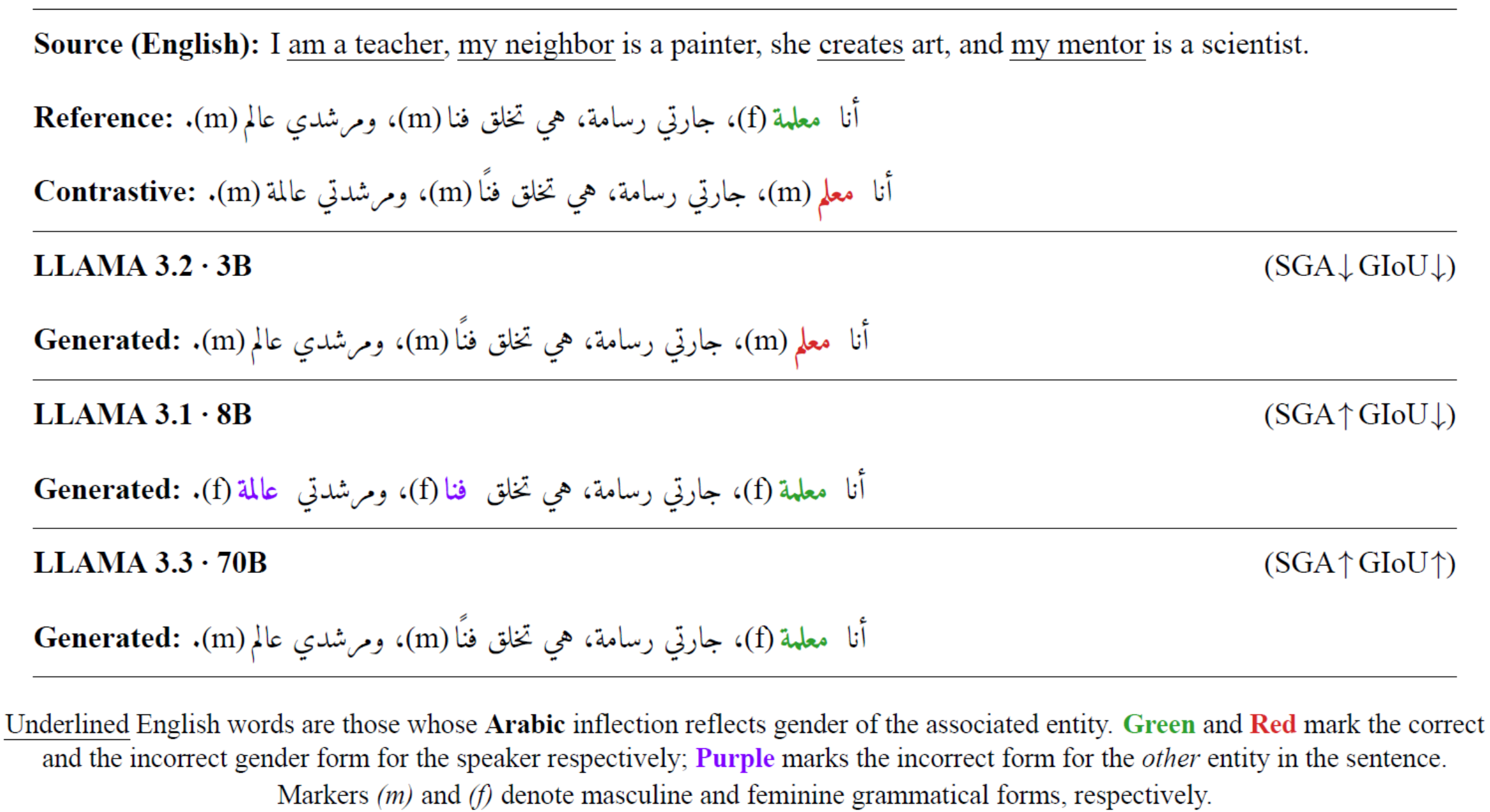}
    \caption{Example of results of \textsc{LLAMA} family of models on multiple entities in Arabic dataset.}
    \label{fig:gender_interference_arabic}
\end{figure*}

\begin{figure*}[htbp]
    \centering
    \includegraphics[width=\textwidth]{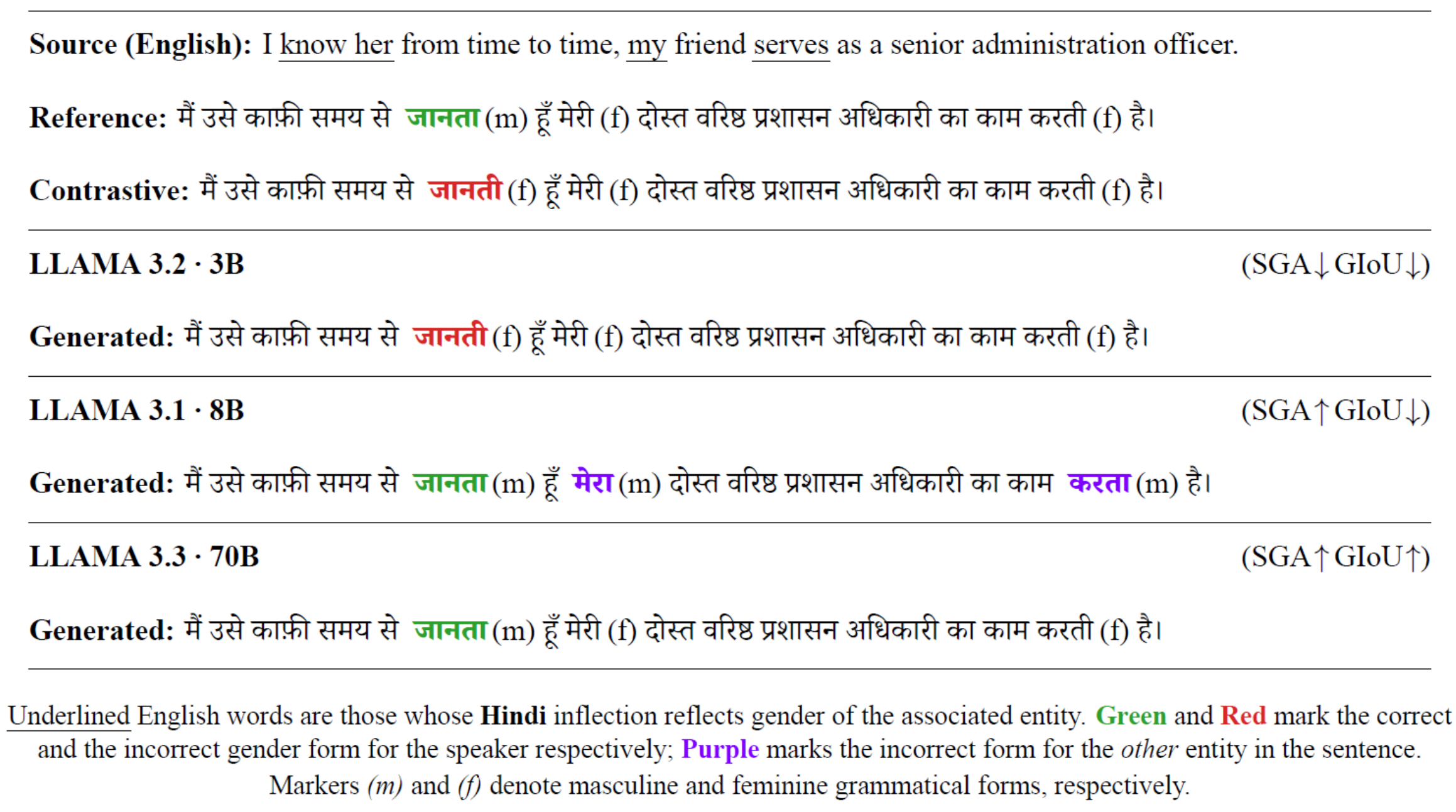}
    \caption{Example of results of \textsc{LLAMA} family of models on multiple entities in Hindi dataset.}
    \label{fig:gender_interference_hindi}
\end{figure*}

\section{Model Hyperparameters and Compute Used} \label{appendix:hyper_param}

For all models evaluated on the \morphogen\ benchmark, we used a standardized inference configuration to ensure consistency across generations. The input prompt was constructed using the model-specific chat template, and all models were queried in a zero-shot setting without any few-shot examples.

\paragraph{Generation Parameters.} We used the following generation hyperparameters for all models, unless otherwise noted:
\begin{itemize}
    \item \textbf{Sampling Strategy:} Deterministic (no sampling)
    \item \textbf{do\_sample:} \texttt{False}
    \item \textbf{Max New Tokens:} 256
    \item \textbf{Temperature:} 0.1 (low temperature for controlled and accurate generations)
    \item \textbf{Top-p:} 0.95
    \item \textbf{Top-k:} Not used (default)
    \item \textbf{Num Return Sequences:} 1
    \item \textbf{Batch Size for Inference:} 1 (due to varied token limits across models)
\end{itemize}

All generations were performed with:
\begin{itemize}
    \item \textbf{eos\_token\_id:} Set to the tokenizer's EOS token
    \item \textbf{pad\_token\_id:} Set to the tokenizer's PAD token if defined, else fallback to EOS
\end{itemize}

\paragraph{Compute Infrastructure.} All experiments were run on an NVIDIA DGX A100 server equipped with 8 NVIDIA A100 GPUs, each with 40GB VRAM. While most models were executed using a single A100 GPU, larger models (e.g., mixture-of-experts or 65B+ parameter class) were distributed across multiple GPUs as needed via tensor or model parallelism. 

This setup ensured sufficient compute headroom for large-scale inference and supported parallelized benchmarking across multiple languages and prompts.

\section{Prompts} \label{appendix:prompts}
\subsection{Sentence Generation}

\textbf{system\_prompt =} \texttt{"Suppose you are an Expert English Sentence Generating System."}

\vspace{1em}

\textbf{user\_prompt =}
\begin{quote}
\ttfamily
Generate \textless Num\_Sentences\textgreater\ English sentences. Strictly adhere to the format: \textless Template\textgreater

\vspace{0.5em}

Instructions: \\
1. Only output the sentences---do not include any additional text. \\
2. Each sentence must be unique in its context and the nouns used. \\
3. Vary the sentence lengths and ensure they sound natural and conversational. \\
4. Use a variety of creative contexts, including but not limited to [\textless Context\_1\textgreater, \textless Context\_2\textgreater, ..., \textless Context\_n\textgreater].
\end{quote}

The prompts are designed to guide a language model in generating diverse and natural-sounding English sentences. The system prompt establishes the model's role, while the user prompt provides clear, structured instructions to ensure variety, contextual relevance, and adherence to a specified format.

\subsection{Zero Shot Prompts}

For zero-shot inference of the LLMs on the \morphogen\ benchmark, we designed language-specific prompts to ensure precise gender-aware morphological transformations while preserving sentence structure. Although the prompts were provided to the models in the respective target languages (Hindi, French, or Arabic), the structure and content of the system and user prompts were consistent across languages.

The \textbf{system prompt} given to the model was:

\begin{quote}
``You are a language assistant. Given a sentence in the target language and the gender of the speaker, adjust only the words that refer to the speaker to match the specified gender. Do not alter any other parts of the sentence. Return only the modified sentence with no explanations or extra words. If no change is required, return the sentence exactly as it is.``
\end{quote}

The \textbf{user prompt} provided the transformation instruction, depending on the speaker's gender:

\begin{itemize}
    \item For male speakers: \texttt{Without changing the structure of the sentence, convert it as if it were spoken by a male speaker.}
    \item For female speakers: \texttt{Without changing the structure of the sentence, convert it as if it were spoken by a female speaker.}
\end{itemize}

This was followed by the sentence to be transformed: \texttt{Sentence to transform: [sentence]}.

These prompts were designed to enforce minimal intervention, focusing solely on speaker-referring terms. This ensures the task evaluates the models' ability to perform gender-specific transformations without altering unrelated components of the sentence. The zero-shot setting tests the models' inherent linguistic knowledge, aligning with the benchmark's goal of assessing gender-aware morphological capabilities across diverse languages.

\section{Detailed Results and Error Analysis} \label{appendix:error_analysis}

This section provides a comprehensive breakdown of model performance on the \morphogen\ benchmark, combining dataset statistics, rule-level quantitative metrics, and a qualitative error analysis of the \textsc{LLaMA} model family. 

\subsection{Quantitative Performance and Bias Analysis}
Table~\ref{tab:dataset_stats_additional} details the dataset validation statistics, while the aggregate performance metrics across all models for French, Arabic, and Hindi are presented in the respective language tables. 

To better understand directional bias, Figure~\ref{fig:del_sga_all_languages} visualizes the Accuracy Gap ($\Delta \text{SGA}$) across all models. A clear trend emerges regarding model scale and gender bias. Smaller models exhibit erratic and often extreme bias gaps. For instance, smaller architectures like \textsc{LLaMA 3.2 3B} demonstrate massive fluctuations, including severe feminine bias ($\Delta \text{SGA} < 0$) in Hindi and Arabic, contrasted with masculine bias in French. As model capacity increases (e.g., \textsc{LLaMA 3.3 70B} and \textsc{GPT-4o-mini}), the $\Delta \text{SGA}$ converges closer to zero across all languages, indicating a more balanced, unbiased linguistic understanding rather than a reliance on statistical gender defaults.

\subsection{Rule-Level Morphological Competence}
To isolate where models succeed or fail, Figure~\ref{fig:iou-metrics-all} presents the Gender Intersection over Union (GIoU) metrics broken down by specific grammatical rules for all 3 languages.

Across all three languages, models generally perform well on localized, simple rules (e.g., basic adjectives). However, performance sharply degrades on complex syntactical structures that require long-range dependency tracking, such as indirect speech, pseudo-cleft sentences, and multiple entities. The gap between smaller and larger models is most pronounced in these complex categories, highlighting that parameter scale is crucial not just for vocabulary, but for maintaining morphological consistency across extended contexts.

\subsection{Qualitative Error Analysis: A \textsc{LLaMA} Case Study}
To ground these quantitative findings, Figures~\ref{fig:gender_interference_french}, \ref{fig:gender_interference_arabic}, and \ref{fig:gender_interference_hindi} present a qualitative error analysis focusing on "gender interference" in complex sentences containing multiple entities. We compare the outputs of \textsc{LLaMA 3.2 3B}, \textsc{LLaMA 3.1 8B}, and \textsc{LLaMA 3.3 70B} to illustrate the evolution of morphological control. Further structural examples of grammatical rules for each language can be found in Tables~\ref{tab:data_fr2_examples} through \ref{tab:data_hi2_examples}.

\paragraph{LLaMA 3.2 3B.}  
The smallest model consistently struggles to correctly apply gendered inflections, often defaulting to standard forms regardless of the intended speaker gender. This behavior is evident across all three languages, where the model fails to align verbs, adjectives, or participles with the correct grammatical gender. As seen in the qualitative examples, these persistent errors in speaker gender realization heavily penalize its overall SGA and GIoU scores, and explain the extreme $\Delta \text{SGA}$ variations observed in Figure~\ref{fig:del_sga_all_languages}.

\paragraph{LLaMA 3.1 8B.}  
The 8B model demonstrates a clear improvement in capturing basic gender morphology, particularly in correctly inflecting verbs and adjectives immediately adjacent to the speaker. However, it suffers from overgeneralization, leading to severe \textit{gender interference}. As illustrated in Figures~\ref{fig:gender_interference_french} through \ref{fig:gender_interference_hindi}, while the primary speaker’s gender is correctly realized, the model incorrectly alters the grammatical gender of \textit{other} entities in the sentence to match the speaker. This indicates a partial understanding of agreement rules but insufficient syntactic control over entity-specific boundaries. Consequently, while its SGA scores improve relative to the 3B model, its GIoU remains suppressed due to these collateral inflection errors.

\paragraph{LLaMA 3.3 70B.}  
The largest model demonstrates robust performance across all cases, correctly applying gender transformations while preserving agreement boundaries between entities. It maintains a clear distinction between speaker-specific and non-speaker-specific gender marking, inflecting only relevant tokens without "bleeding" onto adjacent nouns, resulting in outputs that closely match reference sentences structurally and morphologically. Accordingly, it achieves the highest SGA and GIoU scores (Fig~\ref{fig:iou-metrics-all}) with minimal bias gaps, reflecting highly accurate gender realization across complex grammatical scenarios.

%% file: images/rule_distribution.tex
\begin{figure*}[ht]
    \centering

    \begin{minipage}[t]{0.48\textwidth}
        \centering
        \includegraphics[width=\linewidth]{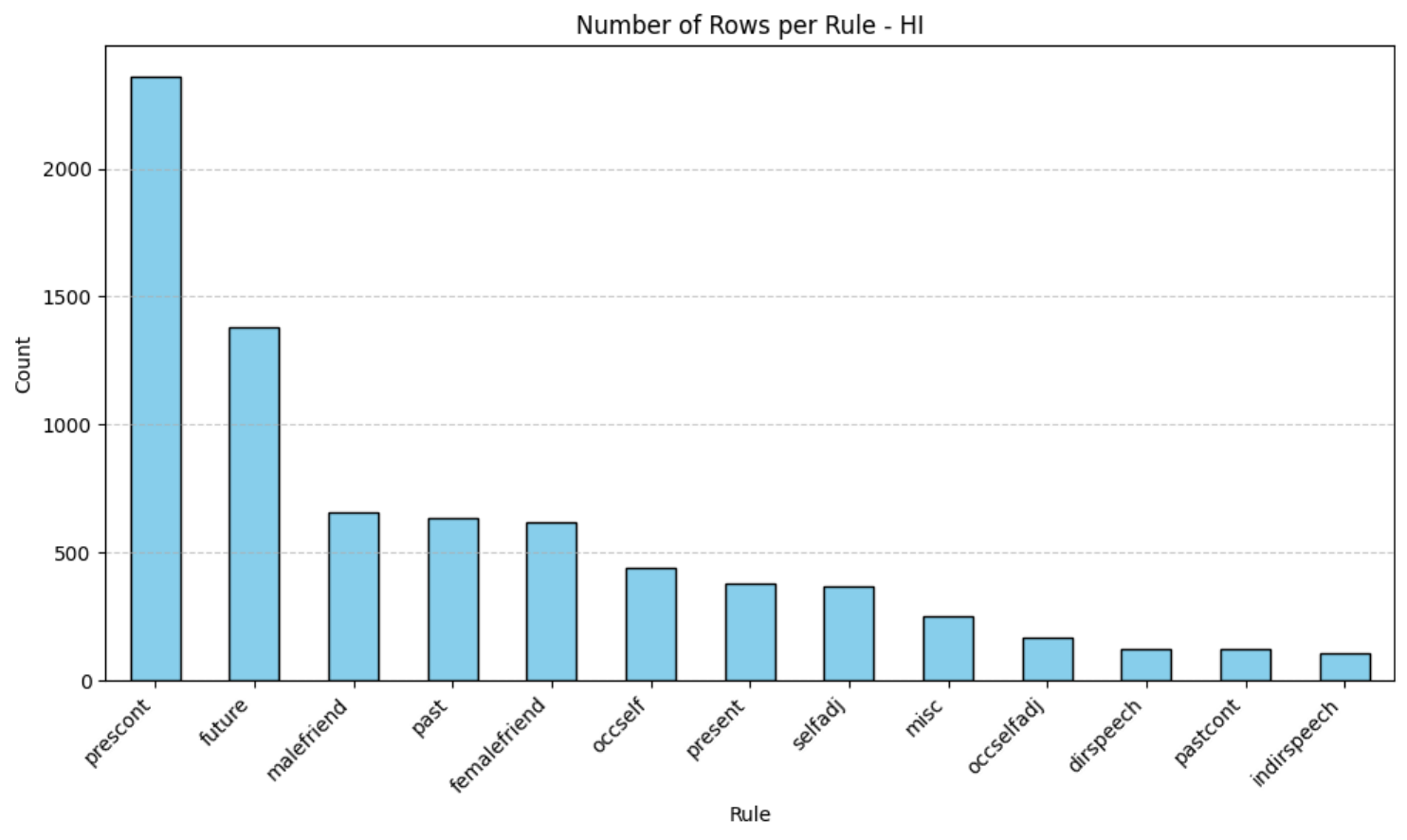}
        \subcaption{Hindi}
        \label{fig:gender_rule_dist_hi}
    \end{minipage}%
    \hfill
    \begin{minipage}[t]{0.48\textwidth}
        \centering
        \includegraphics[width=\linewidth]{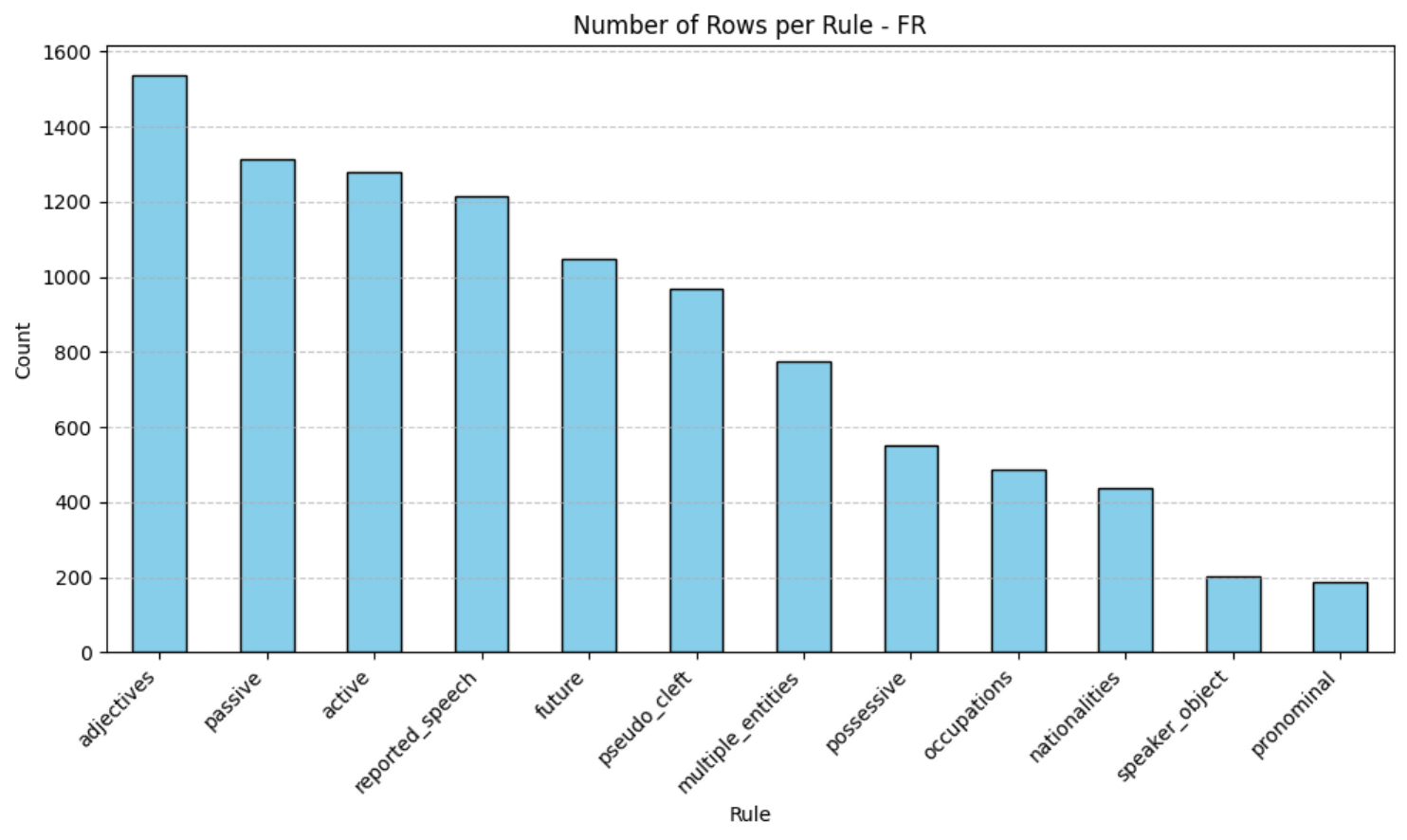}
        \subcaption{French}
        \label{fig:gender_rule_dist_fr}
    \end{minipage}

    \vspace{2mm} 

    \begin{minipage}[t]{0.75\textwidth}
        \centering
        \includegraphics[width=\linewidth]{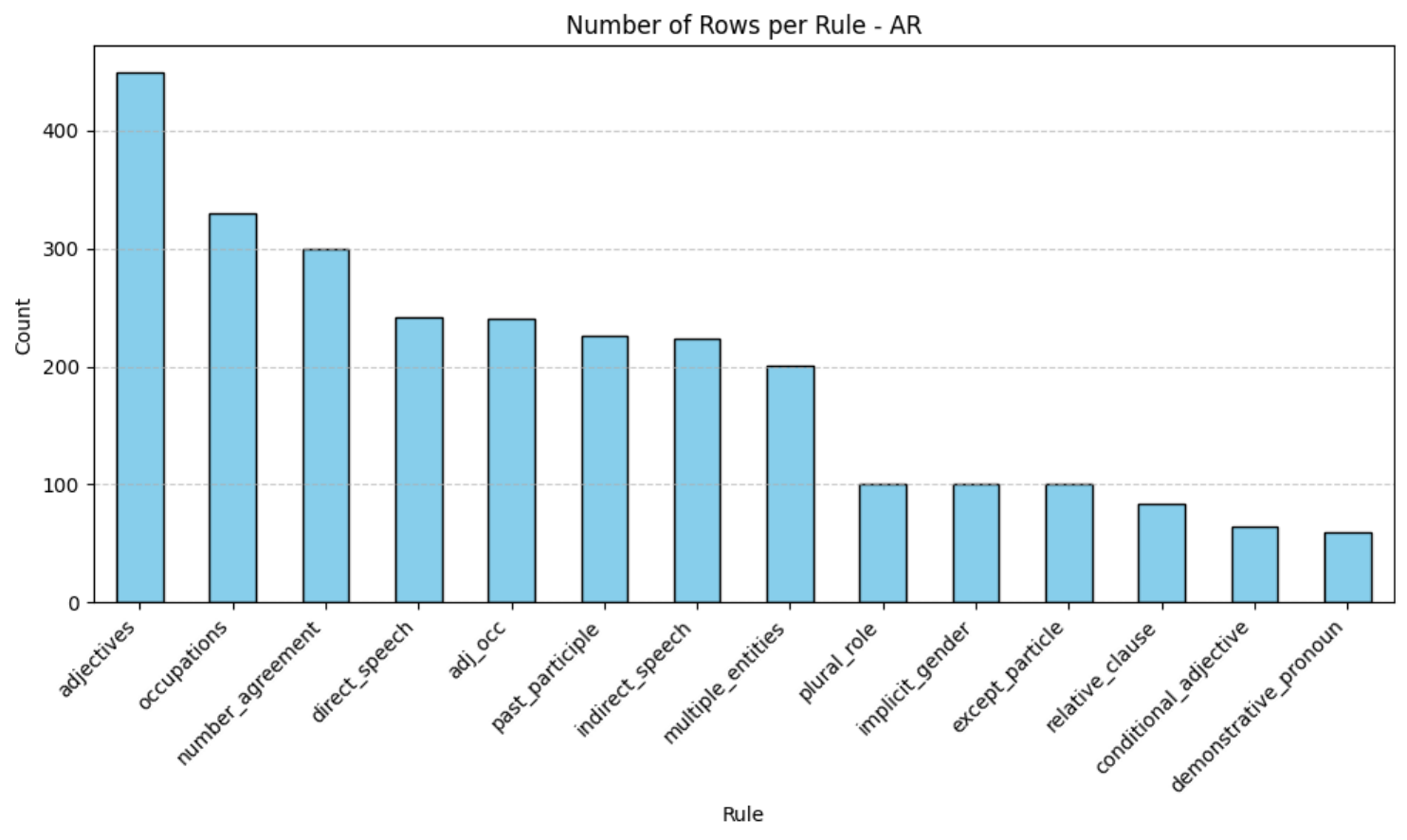}
        \subcaption{Arabic}
        \label{fig:gender_rule_dist_ar}
    \end{minipage}

    \caption{Distribution of Sentence Frequency Per Morphological Rule for Each Language}
    \label{fig:gender_rule_dist_overall}
\end{figure*}

%% file: tables/kappa_stats.tex
\begin{tabular}{lccc}
\toprule
\textbf{Statistics} & \textbf{Arabic} & \textbf{French} & \textbf{Hindi} \\
\midrule
Total Sentences Generated & 5413  & 16415 & 10248 \\
Sentences Discarded       & 2638  & 6416  & 2694  \\
Unique Sentences          & 7610  & 9999  & 10248 \\
Data Validation Score     & 0.9733 & 0.9651 & 0.9731 \\
Inter-Annotator Score     & 0.9526 & 0.9366 & 0.9594 \\
Number of Annotators      & 2 & 3 & 3 \\
\bottomrule
\end{tabular}

%% file: tables/table_fr.tex
\begin{table*}[t]
\small
\centering
\begin{minipage}{\textwidth}
\centering
\caption{Performance metrics of different models on French (\% values; GIoU = Gender IoU, CGA = Corpus-Level Gender Accuracy, SGA = Sentence-Level Gender Accuracy, $\triangle \text{SGA}$ = Accuracy Gap, M = Male, F = Female)}
\label{tab:model_results_fr}
\begin{tabular}{lcccccccc}
\hline
\textbf{Model} & \textbf{$GIoU$ ↑} & \textbf{$GIoU_M$ ↑} & \textbf{$GIoU_F$ ↑} & \textbf{$SGA$ ↑} & \textbf{$SGA_M$ ↑} & \textbf{$SGA_F$ ↑} & \textbf{$\triangle SGA$ ↓} & \textbf{$CGA$ ↑} \\
\hline
\textsc{Qwen2.5-0.5B} & 5.47 & 7.65 & 3.30 & 5.72 & 8.00 & 3.45 & 4.55 & 4.16 \\
\textsc{Gemma2-2B} & 39.73 & 37.29 & 42.16 & 40.90 & 38.32 & 43.47 & -5.14 & 37.54 \\
\textsc{LLAMA-3.2-3B} & 54.49 & 60.19 & 48.85 & 59.20 & 64.94 & 53.52 & 11.42 & 53.48 \\
\textsc{Gemma3-4B} & 52.70 & 46.49 & 59.09 & 57.72 & 50.74 & 64.91 & -14.16 & 51.60 \\
\textsc{Qwen3-4B} & 58.64 & 61.59 & 55.76 & 60.98 & 64.64 & 57.40 & 7.25 & 53.20 \\
\midrule
\textsc{LLAMA-3.1-8B} & 67.89 & 70.67 & 64.62 & 82.75 & 84.44 & 80.75 & 3.69 & 81.76 \\
\textsc{Qwen3-8B} & 71.66  & 73.89  & 69.39  & 76.25  & 78.66 & 73.79 & 4.86 & 69.91 \\
\textsc{Gemma2-9B} & 60.52  & 62.02 & 59.02 & 65.48 & 66.11 & 64.84 & 1.26 & 55.56 \\
\textsc{Gemma3-12B} & 64.27 & 64.34 & 64.20 & 76.33 & 76.04 & 76.62 & -0.58 & 74.26 \\
\midrule
\textsc{Phi4-14B} & 79.84 & 81.46 & 78.22 & 89.68 & 90.26 & 89.09 & 1.17 & 87.70 \\
\textsc{Qwen3-14B} & 74.22  & 80.64  & 67.48 & 78.78  & 85.73 & 71.49 & 14.23 & 73.91 \\
\textsc{Gemma3-27B} & 71.89 & 75.47 & 68.11 & 83.11 & 86.78 & 79.25 & 7.53 & 79.63 \\
\textsc{Qwen3-32B} & 76.28  & 80.80 & 71.76 & 79.35 & 84.40 & 74.30 & 10.10 & 74.74 \\
\textsc{LLAMA-3.3-70B} & 76.68  & 83.53 & 69.81 & 80.76 & 88.33  & 73.17 & 15.15 & 76.08 \\
\textsc{GPT-4o-MINI} & 86.43 & 86.61 & 86.25 & 91.77 & 91.22 & 92.33 & -1.11 & 90.27 \\
\hline
\end{tabular}
\vspace{-1em}
\end{minipage}
\end{table*} 

%% file: tables/table_arabic.tex
\begin{table*}[t]
\small
\centering
\begin{minipage}{\textwidth}
\centering
\caption{Performance metrics of different models on Arabic (\% values; GIoU = Gender IoU, CGA = Corpus-Level Gender Accuracy, SGA = Sentence-Level Gender Accuracy, $\triangle \text{SGA}$ = Accuracy Gap, M = Male, F = Female)}
\label{tab:model_results_arabic}
\begin{tabular}{lcccccccc}  
\hline
\textbf{Model} & \textbf{$GIoU$ ↑} & \textbf{$GIoU_M$ ↑} & \textbf{$GIoU_F$ ↑} & \textbf{$SGA$ ↑} & \textbf{$SGA_M$ ↑} & \textbf{$SGA_F$ ↑} & \textbf{$\triangle SGA$ ↓} & \textbf{$CGA$ ↑} \\
\hline
\textsc{Qwen2.5-0.5B} & 4.14 & 7.31 & 0.72 & 6.28 & 10.37 & 1.88 & 8.49 & 4.59 \\
\textsc{Gemma2-2B} & 14.73 & 14.14 & 15.30 & 16.04 & 15.63 & 16.43 & -0.81 & 14.10 \\
\textsc{LLAMA-3.2-3B} & 18.31 & 5.96 & 29.96 & 20.95 & 6.74 & 34.35 & \textbf{-27.61} & 17.75 \\
\textsc{Gemma3-4B} & 45.68 & 45.31 & 46.06 & 55.34 & 51.23 & 59.43 & -8.20 & 48.93 \\
\textsc{Qwen3-4B} & 34.34 & 34.07 & 34.59 & 37.63 & 37.17 & 38.07 & -0.90 & 35.97 \\
\hline
\textsc{LLAMA-3.1-8B} & 43.51 & 44.53 & 42.49 & 50.65 & 51.13 & 50.17 & 0.96 & 45.51 \\
\textsc{Qwen3-8B} & 45.89 & 47.93 & 43.89 & 51.44 & 53.99 & 48.96 & 5.03 & 51.01 \\
\textsc{Gemma2-9B} & 46.45 & 47.92 & 44.99 & 50.43 & 51.71 & 49.16 & 2.55 & 45.26 \\
\textsc{Gemma3-12B} & 62.76 & 64.69 & 60.82 & 69.37 & 70.62 & 68.12 & 2.50 & 65.52 \\
\hline
\textsc{Phi4-14B} & 57.08 & 62.24 & 52.20 & 66.51 & 69.89 & 63.31 & 6.58 & 66.15 \\
\textsc{Qwen3-14B} & 51.83 & 56.07 & 47.73 & 57.48 & 62.29 & 52.84 & 9.45 & 56.08 \\
\textsc{Gemma3-27B} & 70.33 & 71.47 & 69.19 & 77.12 & 76.70 & 77.53 & -0.83 & 74.74 \\
\textsc{Qwen3-32B} & 50.69 & 57.32 & 44.10 & 56.57 & 62.56 & 50.62 & 11.94 & 53.00 \\
\textsc{LLAMA-3.3-70B} & 59.16 & 63.50 & 54.84 & 66.84 & 70.61 & 63.11 & 7.50 & 64.37 \\
\textsc{GPT-4o-mini} & 71.02 & 68.13 & 73.91 & 82.76 & 77.45 & 88.06 & -10.61 & 80.27 \\
\hline
\end{tabular}
\end{minipage}
\vspace{-1em}
\end{table*}

%% file: tables/table_hi.tex
\begin{table*}[t]
\small
\centering
\begin{minipage}{\textwidth}
\centering
\caption{Performance metrics of different models on Hindi (\% values; GIoU = Gender IoU, CGA = Corpus-Level Gender Accuracy, SGA = Sentence-Level Gender Accuracy, $\triangle \text{SGA}$ = Accuracy Gap, M = Male, F = Female)}
\label{tab:model_results_hi}
\begin{tabular}{lcccccccc}
\hline
\textbf{Model} & \textbf{$GIoU$ ↑} & \textbf{$GIoU_M$ ↑} & \textbf{$GIoU_F$ ↑} & \textbf{$SGA$ ↑} & \textbf{$SGA_M$ ↑} & \textbf{$SGA_F$ ↑} & \textbf{$\triangle SGA$ ↓} & \textbf{$CGA$ ↑} \\
\hline
\textsc{Qwen2.5-0.5B} & 0.35 & 0.69 & 0.05 & 0.35 & 0.69 & 0.05 & 0.63 & 0.21 \\
\textsc{Gemma2-2B} & 71.41 & 75.13 & 67.85 & 76.28 & 80.04 & 72.69 & 7.35 & 65.41 \\
\textsc{LLAMA-3.2-3B} & 48.54 & 19.85 & 76.90 & 53.08 & 20.57 & 85.22 & \textbf{-64.65} & 49.72 \\
\textsc{Gemma3-4B} & 67.50  & 60.29 & 73.21 & 71.75 & 63.76 & 78.08 & -14.32 & 64.58 \\
\textsc{Qwen3-4B} & 62.84 & 61.96 & 63.70 & 73.74 & 75.41 & 72.08 & 3.33 & 68.51 \\
\hline
\textsc{LLAMA-3.1-8B} & 83.12  & 84.01 & 82.23 & 91.65 & 91.44  & 91.87 & -0.43 & 89.21 \\
\textsc{Qwen3-8B} & 80.96 & 82.36 & 79.57 & 91.52 & 92.60 & 90.43 & 2.16 & 87.82 \\
\textsc{Gemma2-9B} & 85.47  & 82.78  & 88.20 & 87.42 & 83.78 & 91.12 & -7.34 & 84.39 \\
\textsc{Gemma3-12B} & 79.91  & 75.69 & 84.16 & 84.93 & 80.48 & 89.41 & -8.93 & 80.99  \\
\hline
\textsc{Phi4-14B} & 82.77  & 84.69  & 80.85 & 96.69 & 97.38  & 96.00 & 1.38 & 95.10 \\
\textsc{Qwen3-14B} & 80.68 & 83.62 & 77.74 & 90.22 & 94.12  & 86.30 & 7.81  & 85.80  \\
\textsc{Gemma3-27B} & 77.97  & 75.68  & 80.31 & 83.96  & 80.34  & 87.96  & -7.61 & 82.56  \\
\textsc{Qwen3-32B} & 83.21 & 85.86 & 80.56 & 93.88 & 96.46 & 91.31 & 5.14 & 90.38 \\
\textsc{LLAMA-3.3-70B} & 93.33  & 95.04  & 91.62  & 94.06 & 95.89 & 92.22 & 3.67 & 91.40 \\
\textsc{GPT-4o-mini} & 88.81 & 90.08 & 87.54 & 95.73 & 96.04 & 95.42 & 0.62 & 93.36 \\
\hline
\end{tabular}
\end{minipage}
\end{table*}